\documentclass{article} 
\usepackage[final]{colm2026_conference}

\usepackage{microtype}
\usepackage{hyperref}
\usepackage{url}
\usepackage{booktabs}

\usepackage{lineno}

\definecolor{darkblue}{rgb}{0, 0, 0.5}
\hypersetup{colorlinks=true, citecolor=darkblue, linkcolor=darkblue, urlcolor=darkblue}

\title{Understanding Calibration and Truncation Error Propagation in Training-Free Low-Rank Compression for LLMs}

\author{%
Mohanad Odema, Gabrielle De Micheli, Dayin Gou,\\%
\textbf{Nilesh Malpeddi, Prathamesh Vaste, Jacob Song}\\%
LG Electronics, North America\\
\texttt{mohanad.odema@lge.com}
}

\usepackage[utf8]{inputenc} 
\usepackage[T1]{fontenc}    
\usepackage{hyperref}       
\usepackage{url}            
\usepackage{booktabs}       
\usepackage{amsfonts}       
\usepackage{nicefrac}       
\usepackage{microtype}      
\usepackage{xcolor}         

\usepackage{multirow}
\usepackage{makecell}
\usepackage{amsmath}
\usepackage{subcaption}
\usepackage{pgfplots}
\usepackage[table]{xcolor}
\usepackage{graphicx}

\usepackage{wrapfig}

\usepackage{enumitem}
\usepackage[ruled,vlined]{algorithm2e}
\usepackage{bm}

\pgfplotsset{compat=1.18}

\definecolor{first}{RGB}{211, 211, 211}
\definecolor{second}{RGB}{238, 238, 238}

\newcommand{\key}[1]{%
  \begingroup
  \setlength{\fboxsep}{0.3pt}%
  \colorbox{first}{\raisebox{0pt}[1.45ex][0.5ex]{#1}}%
  \endgroup
}

\newcommand{\keymath}[1]{%
  \begingroup
  \setlength{\fboxsep}{1.2pt}%
  \colorbox{first}{$\displaystyle #1$}%
  \endgroup
}

\definecolor{darkgreen}{RGB}{0,120,0}

\newcommand{\squishlist}{
 \begin{list}{$\bullet$}
  { \setlength{\itemsep}{0pt}
     \setlength{\parsep}{3pt}
     \setlength{\topsep}{3pt}
     \setlength{\partopsep}{0pt}
     \setlength{\leftmargin}{1.5em}
     \setlength{\labelwidth}{1em}
     \setlength{\labelsep}{0.5em} } }

\newcommand{\squishlisttwo}{
 \begin{list}{$\bullet$}
  { \setlength{\itemsep}{0pt}
     \setlength{\parsep}{0pt}
    \setlength{\topsep}{0pt}
    \setlength{\partopsep}{0pt}
    \setlength{\leftmargin}{2em}
    \setlength{\labelwidth}{1.5em}
    \setlength{\labelsep}{0.5em} } }

\newcommand{\squishend}{
  \end{list}  }

\begin{document}

\ifcolmsubmission
\linenumbers
\fi

\maketitle

\begin{abstract}

Training-free low-rank compression frameworks have been gaining prominence for LLM compression given their effectiveness in reducing model parameter count while maintaining task-level accuracy. However, existing SOTA frameworks share two key limitations:
(1) residual errors in calibration data activations accumulate across layers during compression, causing misalignment between representations simulated at compression time and those experienced at inference;
(2) the assumption that layer importance distribution is preserved post-compression does not hold.
Together, these two effects introduce misalignment in the compression process in relation to the deployed model.
We study these effects and propose a simple, training-free methodology compatible with existing frameworks to mitigate them, comprising: (1) Layer-by-Layer Compression with Calibration Correction; (2) Iterative Compression with Rank Allocation Correction. Implemented atop an existing SOTA decomposition framework, and evaluated on Llama and Qwen3 models across various benchmarks and compression rates, our approach demonstrates up to $\sim$1-2.5 accuracy point improvements over per-weight and joint decomposition baselines on zero-shot tasks.

\end{abstract}
\section{Introduction}\label{sec:intro}

Large Language models (LLMs)~\citep{touvron2023llama, grattafiori2024llama, achiam2023gpt, gemini2023family, yang2025qwen3} have seen remarkable success in powering numerous AI applications (AI assistants, coding agents, translation and question answering). Despite their impressive capabilities, LLMs contain billions of weight parameters (up to 10B and beyond), exhibiting substantial computational and memory demands on the underlying system, posing deployment challenges especially in constrained environments such as on-device or edge settings. 
Addressing this, model compression has gained prominence to reduce model sizes and alleviate their computational demands through pruning~\citep{lecun1989optimal, hassibi1993optimal, xia2022structured,  ashkboos2024slicegpt, frantar2023sparsegpt, ma2023llmpruner}, low-rank decomposition~\citep{mozaffari2025slim, yuan2023asvd, wang2025svdllm, chiang2026uniql, lin2025modegpt, bai2025ressvd,hu2026saessvd}, and quantization~\citep{frantar2023optq, li2025gptaq, liu2025spinquant, zhang2026qronos}. Moreover, recent approaches have moved towards closed-form training-free compression solutions~\citep{lin2025modegpt, chiang2026uniql} to reduce the model size while effectively matching or surpassing the performance of gradient-based compression approaches which require explicit recovery finetuning (RFT) or expensive Fisher calculations~\citep{ma2023llmpruner, ouderaa2024the}.

Specifically, matrix decomposition approaches, which require limited computing resources and do not require backpropagation, have gained traction as new sophisticated techniques to address the accuracy degradation challenges from Singular Value Decomposition (SVD) at high compression rates. Earlier works~\citep{wang2025svdllm, yuan2023asvd} addressing this problem proposed to apply SVD separately to each weight matrix in a model in an activation-aware manner, where 
whitening is applied using input covariance statistics to project the weight to a high-dimensional space of the input distribution before SVD. Though effective, these approaches still relied on RFT to recover full accuracy. Recent approaches MoDeGPT~\citep{lin2025modegpt} and UniQL~\citep{chiang2026uniql} provide high-performant training-free alternatives that address these challenges and surpass the performance of RFT-based compression works. Their approach involves two key techniques: 
(1) \textbf{Joint Matrix decomposition}; decomposing multiple sublayers together as a module, solving the reconstruction loss minimization problem on a module scale rather than a single weight scope. 
(2) \textbf{Global Rank Allocation Strategy}; a global optimization strategy to non-uniformly distribute sparsity allocations (rank ratios) across layers based on a measure of layer importance, with the goal of maximizing the overall sum of importance scores weighted by the parameters in each layer. Though effective, 
these techniques introduce a compression dependency on the pre-processing stage with regards to (1) \textbf{calibration data collection}, and (2) \textbf{layer rank ratios calculation}, both of which introduce compression misalignment effects.

\begin{figure}[!t]
    \centering
    \begin{subfigure}[b]{0.48\textwidth}
        \centering
        \includegraphics[width=\textwidth]{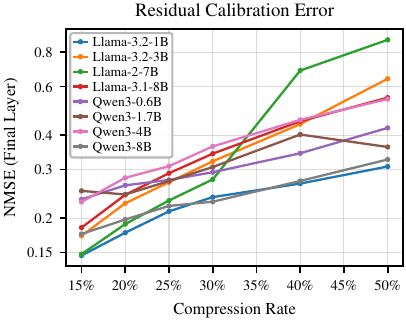}
        \label{fig:calib_a}
    \end{subfigure}
    \begin{subfigure}[b]{0.48\textwidth}
        \centering
        \includegraphics[width=\textwidth]{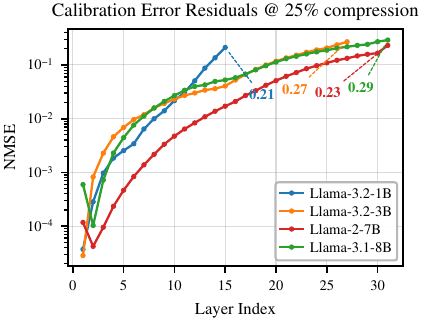} 
        \label{fig:calib_b}
    \end{subfigure}
    \vspace{-2ex}
    \caption{Calibration Error Residual Effects evaluated through NMSE ($\text{NMSE} = \frac{\sum(x - \hat{x})^2}{\sum x^2}$). (\textit{left}) Across different compression rates. (\textit{right}) Per-layer breakdown for Llama models}
    \label{fig:calib_error}
\end{figure}

\textbf{Calibration Error Residuals.}
Low-rank decomposition frameworks typically pass a small subset of input calibration data (e.g., 128 samples from WikiText-2) through the model, and capture calibration data transformations at the input of each weight/module to build covariance statistics. This mechanism fails to account for the residual error that accumulates in calibration data representations as they propagate through the model. Figure \ref{fig:calib_error} illustrates this phenomenon by plotting the per-layer normalized mean square error (NMSE) between the activations of the original model and those of the compressed model at each model layer when compressing through UniQL~\citep{chiang2026uniql}. It can be observed that the errors in the calibration data representations tend to grow as you go deeper in the model. For instance, the Llama-3.2-3B model has NMSE between the calibration activations reaching close to $10^{-2}$ at layer 5 and 0.27 at the final layer, implying a \textbf{growing misalignment between the calibration data representations of the original and compressed models.}

\begin{figure}[!t]
    \centering
    \begin{subfigure}[b]{0.48\textwidth}
        \centering
        \includegraphics[width=\textwidth]{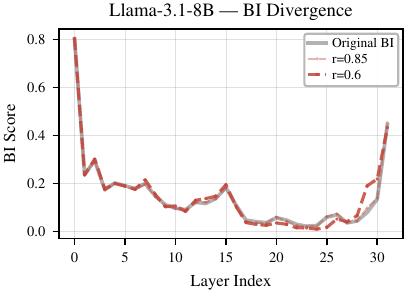}
        \label{fig:div_a}
    \end{subfigure}
    \begin{subfigure}[b]{0.48\textwidth}
        \centering
        \includegraphics[width=\textwidth]{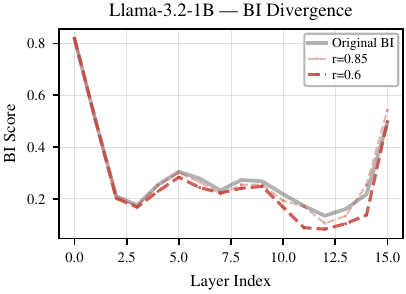} 
        \label{fig:div_b}
    \end{subfigure}
    \vspace{-2ex}
    \caption{BI Score Divergence pre and post compression for Llama-3.1-8B and Llama-3.2-1B. $r$ represents the target rank ratio.}
    \label{fig:div_error}
\end{figure}

\textbf{Rank Ratio Allocation Error Residuals.}
A second source of misalignment arises in the global layer rank allocations, which are precomputed prior to the actual compression based on a measure of layer importance. Figure~\ref{fig:div_error} illustrates this for the Llama-3.1-8B and Llama-3.2-1B models. We reuse the layer importance metric from~\citep{lin2025modegpt, chiang2026uniql} which evaluates a Block Influence (BI) Score~\citep{men2025shortgpt} to measure importance by the degree of transformation between the input and output activations for each layer (more details in Section~\ref{subsec:ratio}). 
As shown in the Figure, the misalignment occurs post-compression where the layer importance scores are liable to drift based on the shifting distributions of weights. For instance, given a rank ratio (1-compression rate) of r=0.85 for the Llama-3.2-1B (right), BI score shifts at layer 12 ($0.13\rightarrow0.10$), and at layer 15 ($0.49\rightarrow0.54$), implying a relative importance change post-compression.

Together, these misalignment effects reveal an underlying limitation: existing training-free low-rank compression frameworks treat the original model as a static reference throughout compression, while in reality each compression step modifies both the calibration signals and the sensitivity estimates on which all subsequent decisions depend. This raises the question of \textit{whether} and \textit{how} to account for these misalignment effects within the compression pipeline, motivating two central questions: \textbf{(1) How should the low-rank reconstruction objectives consider the evolving state of the compressed model?
(2) How should the rank allocation strategy account for the drift in layer importance pre- and post-compression? }

Addressing these RQs, this work makes the following contributions: 

\begin{itemize}
    \item
    We provide an empirical characterization of two sources of misalignment in training-free low-rank compression: the accumulation of residual errors in calibration data activations across layers, and the shift in layer importance estimates before and after compression. We quantify both effects across a range of model families (Llama and Qwen3) and varying compression rates, demonstrating up to 0.27 NMSE in calibration data representations and 0.17 BI score drift in relative layer importance.
    \item 
    We propose two training-free corrections to directly address the aforementioned deficiencies: (1) a layer-per-layer calibration correction that updates the calibration activations used during compression, and (2) an iterative rank allocation correction that refines per-layer truncation ratios to account for the shifted distributions induced by compression. Both methods are directly compatible with existing SOTA decomposition pipelines. 
    \item 
    We evaluate our methods extensively across 8 models from the Llama and Qwen3 families at compression rates of 15\%, 30\%, and 40\%, benchmarking against SOTA per-weight and joint decomposition baselines across five zero-shot downstream tasks. Our approach achieves an overall average gap across all configurations of 0.50pp from the top-performing method, compared to 1.22pp for UniQL and 1.71pp for MoDeGPT. This is >$2\times$ and >$3\times$ closer to the oracle best baseline respectively.
\end{itemize}

\section{Preliminaries}\label{sec:preliminaries}

\subsection{Transformer Basics}

The transformer architecture~\citep{vaswani2017attention} behind LLMs today mainly constitutes two key building blocks: Multi-Head Self Attention (MHSA) and Multi-Layer Perceptron (MLP). These modules contain the bulk of the model parameters. Let $d_{in}$ be the hidden model dimension; $d_{int}$ be the intermediate dimension; $d_h=\frac{d_{in}}{H}$ be the head dimension per attention head. Generally, the output activation from each module can be given as: 
\begin{align}
    f_{\mathrm{MLP}}(X)
    &= \left(\sigma_s(XW_g) \odot XW_u\right)W_d, \\
    f_{\mathrm{MHSA}}(X)
    &= \sum_{i=1}^{H}
    \mathrm{Softmax}\left(
        \frac{
            \rho(XW_q^i)\rho(XW_k^i)^T
        }{\sqrt{d_h}}
    \right)
    (XW_v^i)W_o^i,
\end{align}
where $X\in\mathbb{R}^{N\times d_{in}}$ is the input matrix; $W_q^{i}$, $W_k^{i}$, $W_v^{i}$ are the query, key, and value matrices for head $i$, respectively, with $W_q^{i}$, $W_k^{i}$, $W_v^{i}$ $\in \mathbb{R}^{d_{in} \times d_h}$; $W_g \in \mathbb{R}^{d_{in} \times d_{int}}$, $W_u \in \mathbb{R}^{d_{in} \times d_{int}}$, and $W_d \in \mathbb{R}^{d_{int} \times d_{in}}$ indicate the respective gate, up-projection, and down projection weight matrices; $W_o^{i} \in \mathbb{R}^{d_h \times d_{in}}$ is the output projection matrix corresponding to head $i$; $\sigma_s$ indicates a non-linear activation function (typically SiLU), and $\rho$ indicates the Rotary Positional Embedding (RoPE)~\citep{su2024roformer} for encoding positional information.

\subsection{Low-Rank Decomposition}

LLM parameter reduction can be achieved through Singular Value Decomposition (SVD), where a weight matrix (or a group of them) can be decomposed into singular matrices and singular values that can be further truncated to a desired rank, reducing the number of parameters while minimizing the reconstruction error from the original weight:
\begin{equation}
    W = U\Sigma V^T; \;\; \hat{W} \approx U_r \Sigma_r V^T_r, 
\end{equation}
where $W$ and $\hat{W}$ are the original and truncated weight matrices; $U$ and $V$ are the left and right singular matrices; $\Sigma$ contains the singular values, and $r$ denotes the truncation rank. 

Recently, works have adopted \textbf{Activation-aware} \textbf{decomposition} which employs calibration data, $X$, to optimize the decomposition by minimizing the reconstruction loss between original and compressed output activations rather than the weight matrix, maintaining better accuracy performance at high compression rates, achievable through:

\textbf{Per-Weight Decomposition.} Per-weight methods~\citep{wang2025svdllm, li2025adasvd} approximate each weight matrix to minimize the local per-layer reconstruction loss:
\begin{equation}
    O = min(||WX - \hat{W}X||_2).\label{eqn:single_dec}
\end{equation}
\textbf{Joint Decomposition.} Joint methods~\citep{lin2025modegpt, chiang2026uniql} group related layers within a transformer block as modules, and target minimizing the output activation reconstruction error on the module level from the original and compressed modules:
\begin{equation}
    O = min(f(||X; W_1, \cdots, W_k||_2) - f(||X; \hat{W}_1, \cdots, \hat{W}_k)||)_2, \label{eqn:joint_dec}
\end{equation}
where $\hat{W}_1, \cdots,\hat{W}_k$ indicate the collection of $k$ approximate weights of a module undergoing low-rank compression. This joint decomposition approach has been shown to be superior to per-weight decomposition approaches.

\subsection{Rank Ratio Allocation}\label{subsec:ratio}

Rather than assigning a uniform truncation ratio to all layers, ratios can be assigned non-uniformly based on the importance of each layer. One widely adopted method to gauge the importance of each layer is using the Block Influence (BI) score~\citep{men2025shortgpt}, which quantifies the degree of transformation a layer induces on its input through the complement of cosine similarity (CosSim) between its input and output activations given as 
\begin{equation}BI_i = 1 - \text{CosSim}(X_i, X_{i+1}),
\end{equation}
where $\text{CosSim} = \frac{a \cdot b}{||a||||b||}$ denotes the cosine similarity between two vectors, and $X_i, X_{i+1}$ are the input and output activations of layer $i$, respectively.
A higher BI score indicates a layer induces richer transformations on its inputs, and hence can be assigned a higher importance score to be preserved.
The importance scores can be further diffused across layers using a temperature-scaled Softmax to yield the final rank ratio vector 
\begin{equation}
    \phi = L \times {r_{\mathrm{target}}} \times \text{Softmax}\left(\frac{-\mathrm{scores}}{\tau}\right),
\end{equation}
where $L$ is the number of layers; $r_{target}$ is the overall target sparsity ratio; $scores$ is the vector of per-layer $BI$ scores; and $\tau$ is the temperature parameter. 
The ratio vector $\phi$ must satisfy the global retention budget $r_{\mathrm{target}}$. 

\section{Proposed Method} \label{sec:method}

\begin{figure}[t]
\centering
\begin{minipage}{0.95\columnwidth}
\IncMargin{0.8em}

\begin{algorithm}[H]
\small
\LinesNumbered
\caption{Iterative compression and calibration correction. \key{Key steps shaded.}}
\label{alg:calib}
\KwIn{Model $M$; input calibration $x_{\mathrm{in}}$; target rank ratio $r_{\mathrm{target}}$; dampening coefficient $\alpha$}
\KwOut{Compressed model $\hat{M}_n$}

$\phi \gets \textsc{Compute\_Layer\_Ratios}(M,\ r_{\mathrm{target}})\;$ \tcp*{initialize layer rank ratios}
\For{\key{$n \in \{1, \dots, N\}$}}{ \tcp{N Compression Rounds}
    $x \gets x_{\mathrm{in}}\;$ \tcp*{initialize calibration}
    \For{$l \in \{1, \dots, |M|\}$}{ \tcp{for each model layer}
    $\hat{M}_n[l] \gets \textsc{\textbf{Compress\_Layer}}(M[l],\ x, \phi_l)$\;
    \key{\(x \gets \hat{M}_n[l](x)\)} \tcp*{update calibration for next layer}
}
    \key{$\hat{\phi} \gets \textsc{\textbf{Update\_Layer\_Ratios}}(\hat{M}_n,\ r_{\mathrm{target}},\ \phi,\ \alpha)\;$} \tcp*{update layer rank ratios}
    \eIf{$\textsc{\textbf{Div}}(\phi, \hat{\phi}) < \epsilon$}{ 
        $\text{break}\;$ \tcp*{rank ratios converged}
    }{
        $\phi \gets \hat{\phi}\;$ \tcp*{assign updated rank ratios}
    }
}
\Return $\hat{M}_n$\;
\end{algorithm}

\end{minipage}
\end{figure}

Algorithm \ref{alg:calib} shows our proposed correction framework addressing the aforementioned misalignment factors experienced in training-free low-rank compression pipelines. Lines with shaded background indicate the key novel components: (1) (Line 6) Layer-by-layer correction with calibration error correction; (2) (Lines 2,7) N-round iterative compression with rank allocation correction. Notably, our approaches are designed to be compatible with any training-free low-rank compression framework. We detail the method in the following.

\subsection{Layer-by-Layer Compression with Calibration Correction}

Instead of collecting calibration data transformations in a single pre-processing step across all layers of the original model, the modified approach entails collecting the calibration data in a layer-by-layer fashion, interleaving layer compression operation with calibration updates for the subsequent layers. As seen in Algorithm 1 (line 6), once a layer $l$ is compressed, a forward pass through the newly compressed layer $\hat{M}[l]$ is enacted to provide updated activations to the subsequent layer $l$+1 as its input. This approach takes into consideration preceding layers' compression effects, and aligns the reconstruction loss with the distribution of model representations expected at inference.  Therefore, if we denote activations from the compressed layers as $\hat{X}$, the decomposition objectives from Equations (\ref{eqn:single_dec}) and (\ref{eqn:joint_dec}) become:
\begin{equation}
    O = min(||WX - \hat{W}\keymath{\hat{X}}||_2),\label{eqn:single_dec_ref}
\end{equation}
\begin{equation}
    O = min(f(||X; W_1, \cdots W_k||_2) - f(||\keymath{\hat{X}}; \hat{W}_1, \cdots \hat{W}_k)||)_2. \label{eqn:joint_dec_ref}
\end{equation}

This scheme virtually introduces no additional overheads as it delays calibration data collection at each layer to be on the fly. 
Though intuitive, this calibration error correction has scarcely been applied in the low-rank compression context. Recent Post-Training Quantization (PTQ) methods~\citep{zhang2026qronos, arai2025quantization, liang2026paroquant} introduced a similar correction philosophy. Here, we generalize the concept to be compatible with training-free low-rank decomposition frameworks (details in Appendix \ref{appdx:tf-decomp}).

\subsection{Iterative Compression with Rank Allocation Correction}

The per-layer truncation ratios $\phi$ are precomputed prior to compression based on the importance scores of the original uncompressed model. Post compression, importance ratios are liable to drift from the original estimates as was illustrated in Figure \ref{fig:div_error}. We describe our approach to account for this effect in the following.

\textbf{N-Round Iterative Compression. }We conduct the compression over $N$ rounds (Line 2), effectively rendering an iterative compression with refinement approach. Each round corrects for the layer importance drift, better aligning the compression process with the layer importance scores of the final compressed model. 

\textbf{Refinement and Exit Criteria.} Algorithm \ref{alg:calib} demonstrates the application of $N$-round compression wrapped around the main compression pipeline. Once a compression round is completed, layer rank ratios are recomputed as $\hat{\phi}$ (Line 7) through an update rule defined below. Once convergence is reached (lines 8-11), the iterative compression can conclude, otherwise, it proceeds for another round of iteration up to $N$ rounds.

\textbf{Delta Update Rule. }We take inspiration from Gated Delta Networks~\citep{widrow1988adaptive, schlag2021linear, yang2025gated}, and leverage a delta update rule to blend in the new layer importance ratios with the original estimates. The motivation stems from the fact that each layer possesses unique information and task proficiency that should be preserved to some extent in relation to the original model despite the shifting layer importance distribution. Using the delta rule, refined layer rank ratios for compression iteration $n$ can be given as
\begin{align}
    &\hat{\phi}_i^{n} = \phi_i +  \mathbf{Pr}(\alpha \times \Delta_i^{n}, r_{\mathrm{target}}), \label{eqn:phi} \\
    &s.t.\; \Delta_i^{n} = \frac{\mathrm{BI}_i^{n} - \mathrm{BI}_i}{\mathrm{max}_{l\in [1,L]} (|\mathrm{BI}_l^{n} - \mathrm{BI}_l|)} \forall i \in \{1, \cdots, L\}, \label{eqn:error}
\end{align}
where $\alpha \in [0,1]$ represents a dampening coefficient controlling the update degree; $\Delta_i^{n}$ 
measures the normalized difference in the Block Influence (BI) scores for the $i^{th}$ layer at the $n^{th}$ compression round from the original BI score. $\mathbf{Pr}$ is the projection function that maps the BI scores onto layer rank ratios based on global rank ratio, $r_{\mathrm{target}}$. These rank allocation corrections can be applied independently of the calibration correction.
\section{Experiments}\label{sec:experiments}

\begin{table}[t]
\centering
\resizebox{0.9\columnwidth}{!}{%
\begin{tabular}{cl cccc cccc}
\toprule
\multirow{2}{*}{\shortstack{Compression\\Ratio}} & \multirow{2}{*}{Method} 
  & \multicolumn{4}{c}{Llama} 
  & \multicolumn{4}{c}{Qwen3} \\
\cmidrule(lr){3-6} \cmidrule(lr){7-10}
& & 3.2-1B & 3.2-3B & 3.1-8B & 2-7B & 1.7B & 4B & 4B-I &8B \\
\midrule\midrule
-- & Original (Uncompressed) & 60.69 & 68.12 & 73.98 & 68.86  & 61.93 & 68.68 & 70.90 & 71.87 \\
\midrule\midrule
\multirow{4}{*}{15\%} 
  & SVD-LLM~\citep{wang2025svdllm} & 41.00 & 45.44 & 57.8 & 56.99 &  47.25 & 55.84 &57.41 & 63.13 \\
  & MoDeGPT~\citep{lin2025modegpt}& 54.48 & 61.45 & 66.71 & 65.69 & 56.34 & 63.36 & \textbf{67.27}&  \textbf{67.30} \\
  & UniQL~\citep{chiang2026uniql} & 53.59 & 63.52 &  \textbf{70.60} &  \textbf{66.77}  & 55.10 & 61.60 & 64.74& 66.01 \\
  \cline{2-10}
  & \cellcolor{second}Ours (C)     & \cellcolor{second} 55.06 & \cellcolor{second} \textbf{63.75} & \cellcolor{second}70.54 &\cellcolor{second}  66.61 &\cellcolor{second} 56.61 &  \cellcolor{second}63.81 & \cellcolor{second}66.59 & \cellcolor{second} 66.11 \\
  & \cellcolor{first}Ours (C+R)     &  \cellcolor{first}\textbf{55.48 }& \cellcolor{first} 63.56 & \cellcolor{first} 70.56 & \cellcolor{first}66.47 & \cellcolor{first} \textbf{56.95} &  \cellcolor{first}\textbf{63.95} &  \cellcolor{first}66.50 &\cellcolor{first}66.33 \\
\midrule\midrule
\multirow{4}{*}{30\%} 
  & SVD-LLM~\citep{wang2025svdllm}  & 36.91 & 38.26 & 43.78 & 49.10 & 40.95 & 45.01 & 47.72 & 54.08 \\
  & MoDeGPT~\citep{lin2025modegpt} & 43.89 & 51.73 & 58.15 & 58.72 & \textbf{47.37 }& \textbf{ 52.86} & 54.37 &55.05 \\
  & UniQL~\citep{chiang2026uniql}   & 43.3 & 55.51 &  63.21 &  \textbf{62.24} & 46.63 & 48.47 & 53.21 & 55.66 \\
  \cline{2-10}
  & \cellcolor{second}Ours (C)    & \cellcolor{second} 45.55 & \cellcolor{second} 55.98 & \cellcolor{second}62.72 & \cellcolor{second}62.02 &  \cellcolor{second}45.59 & \cellcolor{second}49.15 & \cellcolor{second}56.50 &  \cellcolor{second}56.17 \\
  & \cellcolor{first}Ours (C+R)     &  \cellcolor{first}\textbf{46.27}  &  \cellcolor{first}\textbf{56.46} & \cellcolor{first} \textbf{63.66} & \cellcolor{first} 62.15 &  \cellcolor{first}46.72 & \cellcolor{first} 49.45 &  \cellcolor{first}\textbf{56.73} & \cellcolor{first}\textbf{56.56 }\\
\midrule\midrule
\multirow{4}{*}{40\%} 
  & SVD-LLM~\citep{wang2025svdllm}  & 36.30 & 36.64 & 39.27 & 43.34 &  38.27 & 38.50 & 40.40 & 43.00 \\
  & MoDeGPT~\citep{lin2025modegpt} & 41.33 & 46.25 & 50.94 & 53.44 &   \textbf{42.31} &  \textbf{46.53} & 47.09 & 46.97 \\
  & UniQL~\citep{chiang2026uniql}   & 41.27 &  48.93 &  56.06 &  56.42 & 41.71 & 42.54 & 47.13 & 47.20 \\
\cline{2-10}
  & \cellcolor{second}Ours (C)     & \cellcolor{second} 42.25 & \cellcolor{second} 50.40 & \cellcolor{second}56.02 &\cellcolor{second} 56.26 &\cellcolor{second} 41.56 & \cellcolor{second}43.02 & \cellcolor{second}48.41 & \cellcolor{second}46.85 \\
  & \cellcolor{first} Ours (C+R)     &  \cellcolor{first}\textbf{42.54}  & \cellcolor{first} \textbf{50.62} & \cellcolor{first} \textbf{56.28} & \cellcolor{first} \textbf{56.63} & \cellcolor{first}41.29 &  \cellcolor{first}43.18 &  \cellcolor{first}\textbf{48.33} & \cellcolor{first}\textbf{47.21} \\ 
\bottomrule
\end{tabular}%
}
\caption{Average Training-free Accuracy across the five 0-shot benchmarks from LM-Eval as detailed in the experimental setup. The strong and light shades of grey indicate first and second, respectively. \textbf{(C) implies calibration correction; (R) implies layer rank correction}}
\label{tab:zero-shot}
\end{table}

\subsection{Experimental Setup}

\textbf{Models.} We use Llama-\{3.1-8B; 3.2-3B; 3.2-1B\}~\citep{grattafiori2024llama, meta2024llama32}, Llama-2-7B~\citep{touvron2023llama}, and Qwen3-\{8B; 4B; 4B-Instruct-2507; 1.7B\}~\citep{yang2025qwen3}. For brevity, Qwen3-4B-Instruct-2507 can be abbreviated by Qwen3-4B-I or Qwen3-4B-Inst.

\textbf{Baselines.} We compare against training-free decomposition methods SVD-LLM~\citep{wang2025svdllm}, MoDeGPT~\citep{lin2025modegpt}, and UniQL~\citep{chiang2026uniql}.

\textbf{Experimental Setting.} We implement our method on top of the open-source UniQL implementation~\citep{chiang2026uniql} -- the current SOTA in joint decomposition to the best of our knowledge.
We follow the same default experimental setting UniQL employs for a fair comparison: 128 Wikitext2 samples for layer ratios estimation; 128 alpaca samples for layer sorting and compression. All evaluations are in BF16 on NVIDIA RTX 6000 Ada GPUs with 48 GB each. More experimental setup details are given in Appendix \ref{appdx:exp}.  

\textbf{Evaluation Benchmarks.} We follow the UniQL convention and conduct our main evaluations on five 0-shot down stream tasks from LM-Eval Harness~\citep{eval-harness}:  PiQA, Winogrande, Arc-Easy, Arc-Challenge, Hellaswag. By default, we report the accuracy except for Arc-Challenge and Hellaswag for which we report the normalized accuracy. Additional results and analysis including 5-shot MMLU benchmark results are reported in Appendix~\ref{appdx:exps}.

\subsection{Overall Results}

We assess the efficacy of calibration data correction and layer rank ratio correction over the average LM-Eval 0-shot task accuracy performance, comparing our method to the baselines outlined above at compression ratios 15\%, 30\%, and 40\%. The full experimental results are reported in Table~\ref{tab:zero-shot}, revealing that our combined methods (C+R) can lead to average task accuracy improvement reaching $\sim$1-2.5 percentage points (pp). Moreover, we find that our method achieves the following Win (W) / Tie (T) / Loss (L) counts (1) 18 / 3 / 3 vs. UniQL; (2) 18 / 0 / 6 vs. MoDeGPT. We consider a difference in score below 0.05 pp to be a tie.

We observe that the effectiveness of our methods is dependent on multiple factors: 
\textbf{(1) Model Family.} Llama-3.2 models benefit the most from the corrections, unlike the smallest Qwen3 models-\{1.7B, 4B\} which are more sensitive; \textbf{(2) Model Size.} Within each family, models of different sizes exhibit varying affinities (e.g., Qwen3-8B benefits from corrections at 30\% (55.66$\rightarrow$56.56) in contrast to the smallest Qwen3 models). \textbf{(3) Compression Rate.}  At aggressive 40\% compression, our method with all corrections offers the best scores across all Llama models. 
We follow this first analysis with a more detailed investigation below.

\begin{figure}[!t]
    \centering
    \includegraphics[width=0.26\columnwidth]{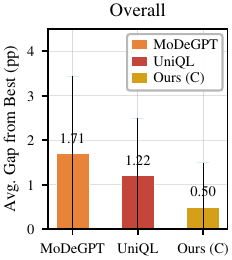}
    \hfill
    \includegraphics[width=0.32\columnwidth]{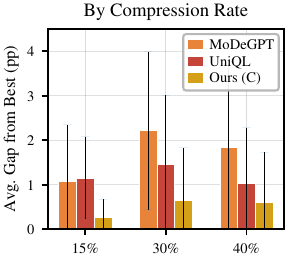}
    \hfill
    \includegraphics[width=0.32\columnwidth]{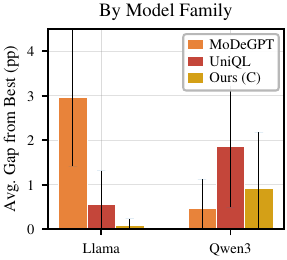}
    \caption{Average Gap from the Best for the different compression methods (Lower is better). Ours here indicates after calibration correction (C). }
    \label{fig:avg_gap}
\end{figure}

\subsection{Calibration Representation Correction (C)}

We further analyze accuracy improvements through quantifying \textbf{Average Gap from Best} as a comparison metric to estimate how far each method is in pp from the best value achieved by any method (i.e., lower is better). In Figure \ref{fig:avg_gap}, we focus on the calibration correction (C) and compare its Average Gap metric against those of MoDeGPT and UniQL from three perspectives: \textit{Overall}, \textit{By Compression Rate}, and \textit{By Model Family}, discussed as follows:

\textbf{(1) Overall / By Compression Rate.} Our corrective calibration achieves the \textbf{smallest (best)} gap from the best score on average, 
at least closer to the top performing solution by a factor of $\sim2\times$ than the next best baseline in the overall.

\textbf{(2) By Model Family.} For the Llama family, our corrective calibration achieves the smallest (best) average gap from the best score. For the Qwen3 family, MoDeGPT achieves the smallest gap. Interestingly, MoDeGPT's average gap from the best for the Llama family is 2.9 pp, while our corrective calibration's average gap for the Qwen3 family is 0.9 pp, implying a more stable performance as observed in the overall case. 

\begin{wraptable}{r}{0.4\textwidth}
\vspace{-2ex}
\centering
\resizebox{0.38\textwidth}{!}{%
\begin{tabular}{lcc}
\toprule
Method & Llama-3.2-1B & Llama-3.1-8B \\
\midrule
MoDeGPT & 0h20m & 3h1m  \\
UniQL   & 0h5m  & 0h19m \\
Ours (C)    & 0h8m  & 0h35m \\
\bottomrule
\end{tabular}%
}
\caption{Compression Wall clock time measured on RTX Ada 6000}
\label{tab:wallclock}
\vspace{-2ex}
\end{wraptable}

\textbf{Compression Quality vs. Complexity Trade-off.} Table \ref{tab:wallclock} shows the compression wall clock time experienced by each method. The table underpins MoDeGPT's inherent complexity and solution quality in providing an accurate solution to the modular matrix reconstruction objective by using expensive pseudo-inverse operations. Our calibration correction solution is built on top of UniQL presenting a more efficient, pseudo-inverse-free alternative, making it $\sim2.5-5\times$ faster than MoDeGPT.

\subsection{Rank Allocation Correction (R)}

\begin{figure}[!t]
    \centering
    \includegraphics[width=0.32\columnwidth]{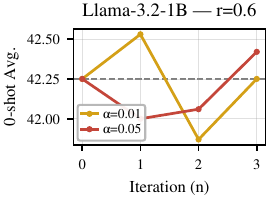}
    \includegraphics[width=0.32\columnwidth]{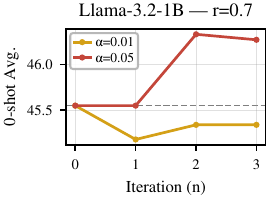}
    \includegraphics[width=0.32\columnwidth]{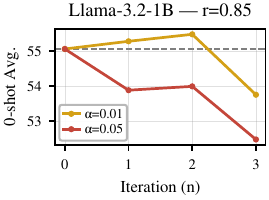}
    \vspace{-2ex}
    \caption{Avg. 0-shot accuracy for Llama-3.2-1B post BI-scores correction (Higher is better).}
    \label{fig:1b_bi_iter}
\end{figure}

Unless otherwise stated, we stack rank allocation correction on top of the calibration correction and assess the (C+R) performance improvement. We experiment with dampening coefficients $\alpha\in\{0.01, 0.05\}$ and $N=3$ rounds, analyzing how the accuracy changes compared to calibration correction (C) (the horizontal dashed lines in Figure~\ref{fig:1b_bi_iter}). We observe the choice of $\alpha$ and $N$ is quintessential to maximizing performance improvement gains over the baselines. For instance, Figure \ref{fig:1b_bi_iter} shows that for Llama-3.2-1B: (1) for $r=0.6$, the parameters $\alpha=0.01$ and $n=1$ yield the largest accuracy improvement by +0.29 pp; (2) for $r=0.7$, the parameters $\alpha=0.05$ and $n=3$ lead to higher accuracy improvement by +0.72. 

Based on these observations, the analysis provides two key takeaways: (1) The choice of best hyperparameters for the delta update rule is dependent on the model type, size, layer importance distribution, and compression rate; (2) a small number of rounds, $N\in[1,3]$ can be sufficient to enhance the average task accuracy.
More experiments are in Appendix~\ref{appdx:exps}.

\begin{figure}[!t]
    \centering
    \includegraphics[width=0.32\columnwidth]{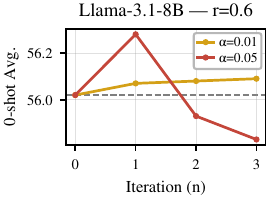}
    \includegraphics[width=0.32\columnwidth]{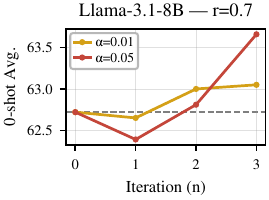}
    \includegraphics[width=0.32\columnwidth]{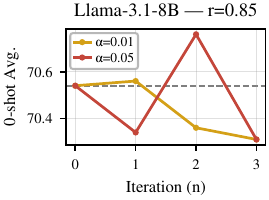}
    \vspace{-2ex}
    \caption{Avg. 0-shot accuracy for Llama-3.1-8B post BI scores correction (Higher is better). }
    \label{fig:8b_bi_iter}
\end{figure}

\subsection{Ablation Studies}\label{subsec:ablation}

\textbf{Repeated Runs.} We repeat our experiments using 3 additional randomly sampled seeds at 40\% compression of Llama models in Table \ref{tab:seeds}, demonstrating that our findings hold.

\begin{table}[t]
\footnotesize
\centering
\begin{tabular}{llcc}
\toprule
\textbf{Model} & \textbf{Config} & \textbf{WikiText-2 PPL $\downarrow$} & \textbf{0-shot Avg. $\uparrow$} \\
\midrule
\multirow{3}{*}{Llama 3.2-1B}
& UniQL                 & $223.01 \pm 12.25$ & $41.78 \pm 0.48$ \\
& \cellcolor{second} Ours (C)              & \cellcolor{second} $218.37 \pm 10.50$ & \cellcolor{second} $41.80 \pm 0.53$ \\
& \cellcolor{first} Ours (C+R)@$N=1$      & \cellcolor{first} $222.79 \pm 0.00$  & \cellcolor{first} $41.93 \pm 0.55$ \\
\midrule
\multirow{3}{*}{Llama 3.2-3B}
& UniQL                 & $70.85 \pm 1.28$ & $49.40 \pm 0.63$ \\
& \cellcolor{second} Ours (C)              & \cellcolor{second} $62.52 \pm 1.13$ & \cellcolor{second} $50.10 \pm 0.52$ \\
& \cellcolor{first} Ours (C+R)@$N=1$      & \cellcolor{first} $60.68 \pm 3.89$ & \cellcolor{first} $50.17 \pm 0.45$ \\
\bottomrule
\end{tabular}
\caption{Results variance after repeated runs with 3 different seeds at 40\% compression. }
\label{tab:seeds}
\end{table}

\textbf{Scaling to 32B Parameter Model.}
We extend our evaluation to a Qwen2.5-32B targeting 40\% compression, applying both our corrective calibration (C) and iterative refinement (R). This experiment is conducted on a single NVIDIA RTX PRO 6000 Blackwell GPU with 96 GB of memory. The results are shown in Table \ref{tab:qwen32b}. 
We observe that unlike the smaller models, pure corrective calibration (C) does not lead to performance improvements across the different benchmarks. Thus, we focus the rest of the experimentation on applying the iterative refinement (R) directly onto the UniQL base, to which we identify performance improvement opportunities. Specifically, Ours (R) at $\alpha$=0.01 N=2 improves 0-shot avg score by 0.3 pp and MMLU by 0.05. This experiment demonstrates larger models are generally more stable and resilient to calibration drifts and layer sensitivity shifts.

\textbf{Additional Ablations.} Ablations on $N$ and $\alpha$ effects are in Appendix \ref{appdx:exps}. Ablation on the compatibility with post-hoc correction methods (EoRA~\citep{liu2024eora}) in Appendix \ref{appdx:eora}.

\begin{table*}[t]
\centering
\footnotesize
\setlength{\tabcolsep}{8pt}
\renewcommand{\arraystretch}{1.12}
\begin{tabular}{@{}lccccc@{}}
\toprule
\textbf{Method}
& \boldmath$\alpha$
& \boldmath$N$
& \textbf{WikiText-2 PPL} $\downarrow$
& \textbf{0-shot Avg.} $\uparrow$
& \textbf{MMLU} $\uparrow$ \\
\midrule

UniQL (base)
& -- & --
& 20.50
& 59.84
& 32.40 \\
\midrule
Ours (C)
& -- & --
& 24.16
& 57.42
& 27.20 \\

\midrule

\multirow{4}{*}{Ours (R)}
& 0.01 & 1
& 20.67
& 59.90
& 32.40 \\

& 0.01 & 2
& 20.69
& \textbf{60.14}
& 32.45 \\

& 0.01 & 3
& 20.70
& 59.88
& 32.46 \\

& 0.05 & 1
& \textbf{20.48}
& 59.54
& \textbf{32.60} \\

\bottomrule
\end{tabular}
\caption{
Results for Qwen2.5-32B at 40\% compression.
Bold indicates the best result.
}
\label{tab:qwen32b}
\end{table*}
\section{Related Works}

\textbf{LLM Weights Compression. }Pruning~\citep{ma2023llmpruner, ashkboos2024slicegpt, frantar2023sparsegpt, sengupta2025you, li2025trthepruner, yang2025let, guo2025slimllm, men2025shortgpt, cai2025llamaflex, saukh2026cut, le2025probe} and Low-rank Approximation~\citep{yuan2023asvd, wang2025svdllm, lin2025modegpt, chiang2026uniql, hu2026saessvd, bai2025ressvd, mozaffari2025slim, koike2025latentllm, wang2025basis, li2025moesvd, huang2025sola, wang2025qsvd} represent effective parameter reduction techniques for reducing memory footprint and computational FLOPs via removing redundant weights or decomposing weight matrices into low-rank structures.

\textbf{Training-Free Parameter Reduction. } This branch of LLM compression works relies on leveraging training-free compression to align the model compression process with the input data distribution. SliceGPT~\citep{ashkboos2024slicegpt} is a notable work leveraging orthogonal matrix transformations to replace each weight with a smaller dense matrix, reducing the embedding dimensions throughout the model. SliceGPT laid the foundation for subsequent training-free matrix decomposition: (1) per-weight~\citep{wang2025svdllm, yuan2023asvd}; (2) per-module (joint decomposition)~\citep{lin2025modegpt, chiang2026uniql, wang2025basis}.

\textbf{Compression Error Correction. }Error correction in LLM compression has been  largely adopted for pruning~\citep{frantar2023sparsegpt, guo2025optimal, liu2024eora}, Post-Training Quantization (PTQ)~\citep{frantar2023optq, li2025gptaq, arai2025quantization, zhang2026qronos, liang2026paroquant, li2025svdquant}, and SVD-based compression~\citep{bai2025ressvd, hu2026saessvd}. 
The work by~\citep{hu2026saessvd} also considers inter-layer error compensation for per-weight SVD-truncation baselines. From the PTQ literature, there is no clear consensus on the corrective calibration correction, specifically on whether the original or compressed representation should be used to guide the quantization process: methods in~\citep{frantar2023optq, zhang2026qronos, liang2026paroquant} work on the compressed representations, whereas methods in~\citep{li2025gptaq, chee2023quip} leverage the original ones. This work departs from others in that it studies these effects for state-of-the-art joint training-free decomposition frameworks, and investigates an additional source of misalignment in layer rank ratios allocation. 

\textbf{Rank Ratio Allocation Strategies.} Early low-rank compression works assigned the target truncation ratio uniformly across layers~\citep{li2025adasvd, wang2025svdllm}.  As LLM layers exhibit varying degrees of sensitivity, recent works~\citep{li2025adasvd, lin2025modegpt, chiang2026uniql, hinostroza2026rethinking} shifted towards assigning ranks non-uniformly across layers based on a measure of sensitivity. For instance,  ShortGPT~\citep{men2025shortgpt} defined the Block Influence (BI) score to characterize layer importance based on cosine similarity. 
\section{Discussion}\label{sec:discussion}

Our experiments reveal that the efficacy of the proposed correction mechanisms is affected by multiple factors, discussed in the following: 

\textbf{Model Architecture Effects. }The divergence in performance can be traced to choices pertaining to variations in architecture, size, and training (data and recipe). For example, the Qwen3 models are designed with a deeper and narrower architecture than the Llama models. The former were developed with the intent to function as hybrid instruct/thinking models and operate with a larger vocabulary size. All these factors lead the Qwen3 models to exhibit sensitivity profiles different from those of the Llama models, as seen in Figures \ref{fig:calib_llamas}-\ref{fig:div_error_3}. We analyze these effects further in Appendix \ref{appdx:exps}.

\textbf{Calibration Data Effects. } Calibration data play a key role in estimation of covariance statistics needed for training-free compression. Recent studies~\citep{ji2025beware, williams-aletras-2024-impact, he2025preserving} have shown that compression calibration data sampled from the same distribution domain as the pretraining data aids the model in retaining its performance more effectively post compression. This can explain the more effective compression of some models when compressed via certain calibration datasets.

\textbf{Hyperparameter Choices. }In our corrective compression, the choice of $\alpha$ and $N$ for our update rule is analogous to adjusting learning rate and number of epochs in finetuning, making techniques like learning rate schedulers relevant to $\alpha$ choices. Additionally, in order to effectively navigate the hyperparameter space, low-cost evaluation mechanisms are needed to gauge if further compression rounds are needed. One manifestation here is the divergence comparison we have in Algorithm \ref{alg:calib} (Lines 8-11) which decides based on the degree of divergence whether another round is needed. Future work can investigate further along these directions.

\textbf{Importance of Pooled Comparisons.} Given all these variations, this motivates the aggregate cross-condition analysis to capture holistic perspectives on the strengths and weaknesses of each technique. Specifically, our Average Gap from Best analysis assesses how each compression method performs across models and category groupings relative to the best result, highlighting the corrective approach's merit in maintaining the least overall gap to best of 0.50 pp versus 1.22 and 1.71 pp for the baselines UniQL and MoDeGPT, respectively.
\section{Conclusion}

In this work, we showed that training-free low-rank compression can become increasingly misaligned as compression progresses, both through residual representation error accumulation in calibration representations and through drift in layer sensitivity estimates. We introduced calibration and layer allocation corrections that address these effects while remaining compatible with existing training-free decomposition frameworks. Across various models and compression rates, the proposed corrections improve the accuracy--compression trade-off and extend the Pareto frontier.

\bibliography{ref}

@misc{eval-harness,
  author       = {Gao, Leo and Tow, Jonathan and Abbasi, Baber and Biderman, Stella and Black, Sid and DiPofi, Anthony and Foster, Charles and Golding, Laurence and Hsu, Jeffrey and Le Noac'h, Alain and Li, Haonan and McDonell, Kyle and Muennighoff, Niklas and Ociepa, Chris and Phang, Jason and Reynolds, Laria and Schoelkopf, Hailey and Skowron, Aviya and Sutawika, Lintang and Tang, Eric and Thite, Anish and Wang, Ben and Wang, Kevin and Zou, Andy},
  title        = {A framework for few-shot language model evaluation},
  month        = 12,
  year         = 2023,
  publisher    = {Zenodo},
  version      = {v0.4.0},
  doi          = {10.5281/zenodo.10256836},
  url          = {https://zenodo.org/records/10256836}
}

@article{grattafiori2024llama,
  title={The llama 3 herd of models},
  author={Grattafiori, Aaron and Dubey, Abhimanyu and Jauhri, Abhinav and Pandey, Abhinav and Kadian, Abhishek and Al-Dahle, Ahmad and Letman, Aiesha and Mathur, Akhil and Schelten, Alan and Vaughan, Alex and others},
  journal={arXiv preprint arXiv:2407.21783},
  year={2024}
}

@article{touvron2023llama,
  title={Llama 2: Open foundation and fine-tuned chat models},
  author={Touvron, Hugo and Martin, Louis and Stone, Kevin and Albert, Peter and Almahairi, Amjad and Babaei, Yasmine and Bashlykov, Nikolay and Batra, Soumya and Bhargava, Prajjwal and Bhosale, Shruti and others},
  journal={arXiv preprint arXiv:2307.09288},
  year={2023}
}

@article{achiam2023gpt,
  title={Gpt-4 technical report},
  author={Achiam, Josh and Adler, Steven and Agarwal, Sandhini and Ahmad, Lama and Akkaya, Ilge and Aleman, Florencia Leoni and Almeida, Diogo and Altenschmidt, Janko and Altman, Sam and Anadkat, Shyamal and others},
  journal={arXiv preprint arXiv:2303.08774},
  year={2023}
}

@article{gemini2023family,
  title       = {Gemini: A Family of Highly Capable Multimodal Models},
  author      = {{Gemini Team, Google DeepMind}},
  journal     = {arXiv preprint arXiv:2312.11805},
  year        = {2023},
}

@article{yang2025qwen3,
  title={Qwen3 technical report},
  author={Yang, An and Li, Anfeng and Yang, Baosong and Zhang, Beichen and Hui, Binyuan and Zheng, Bo and Yu, Bowen and Gao, Chang and Huang, Chengen and Lv, Chenxu and others},
  journal={arXiv preprint arXiv:2505.09388},
  year={2025}
}

@article{vaswani2017attention,
  title={Attention is all you need},
  author={Vaswani, Ashish and Shazeer, Noam and Parmar, Niki and Uszkoreit, Jakob and Jones, Llion and Gomez, Aidan N and Kaiser, {\L}ukasz and Polosukhin, Illia},
  journal={Advances in neural information processing systems},
  volume={30},
  year={2017}
}

@article{su2024roformer,
  title={Roformer: Enhanced transformer with rotary position embedding},
  author={Su, Jianlin and Ahmed, Murtadha and Lu, Yu and Pan, Shengfeng and Bo, Wen and Liu, Yunfeng},
  journal={Neurocomputing},
  volume={568},
  pages={127063},
  year={2024},
  publisher={Elsevier}
}

@incollection{widrow1988adaptive,
  title={Adaptive switching circuits},
  author={Widrow, Bernard and Hoff, Marcian E},
  booktitle={Neurocomputing: foundations of research},
  pages={123--134},
  year={1988}
}

@inproceedings{schlag2021linear,
  title={Linear transformers are secretly fast weight programmers},
  author={Schlag, Imanol and Irie, Kazuki and Schmidhuber, J{\"u}rgen},
  booktitle={International conference on machine learning},
  pages={9355--9366},
  year={2021},
  organization={PMLR}
}

@inproceedings{
yang2025gated,
title={Gated Delta Networks: Improving Mamba2 with Delta Rule},
author={Songlin Yang and Jan Kautz and Ali Hatamizadeh},
booktitle={The Thirteenth International Conference on Learning Representations},
year={2025},
url={https://openreview.net/forum?id=r8H7xhYPwz}
}

@misc{meta2024llama32,
  title        = {LLaMA 3.2 Model Card},
  author       = {Meta AI},
  year         = {2024},
  howpublished = {\url{https://www.llama.com/docs/model-cards-and-prompt-formats/llama3_2/}},
}

@inproceedings{xia2022structured,
  title={Structured pruning learns compact and accurate models},
  author={Xia, Mengzhou and Zhong, Zexuan and Chen, Danqi},
  booktitle={Proceedings of the 60th Annual Meeting of the Association for Computational Linguistics (Volume 1: Long Papers)},
  pages={1513--1528},
  year={2022}
}

@inproceedings{
ma2023llmpruner,
title={{LLM}-Pruner: On the Structural Pruning of Large Language Models},
author={Xinyin Ma and Gongfan Fang and Xinchao Wang},
booktitle={Thirty-seventh Conference on Neural Information Processing Systems},
year={2023},
url={https://openreview.net/forum?id=J8Ajf9WfXP}
}

@inproceedings{
ouderaa2024the,
title={The {LLM} Surgeon},
author={Tycho F. A. van der Ouderaa and Markus Nagel and Mart Van Baalen and Tijmen Blankevoort},
booktitle={The Twelfth International Conference on Learning Representations},
year={2024},
url={https://openreview.net/forum?id=DYIIRgwg2i}
}

@inproceedings{
ashkboos2024slicegpt,
title={Slice{GPT}: Compress Large Language Models by Deleting Rows and Columns},
author={Saleh Ashkboos and Maximilian L. Croci and Marcelo Gennari do Nascimento and Torsten Hoefler and James Hensman},
booktitle={The Twelfth International Conference on Learning Representations},
year={2024},
url={https://openreview.net/forum?id=vXxardq6db}
}

@inproceedings{hassibi1993optimal,
  title={Optimal brain surgeon and general network pruning},
  author={Hassibi, Babak and Stork, David G and Wolff, Gregory J},
  booktitle={IEEE international conference on neural networks},
  pages={293--299},
  year={1993},
  organization={IEEE}
}

@article{lecun1989optimal,
  title={Optimal brain damage},
  author={LeCun, Yann and Denker, John and Solla, Sara},
  journal={Advances in neural information processing systems},
  volume={2},
  year={1989}
}

@inproceedings{frantar2023sparsegpt,
  title={Sparsegpt: Massive language models can be accurately pruned in one-shot},
  author={Frantar, Elias and Alistarh, Dan},
  booktitle={International conference on machine learning},
  pages={10323--10337},
  year={2023},
  organization={PMLR}
}

@article{guo2025optimal,
  title={Optimal brain restoration for joint quantization and sparsification of llms},
  author={Guo, Hang and Li, Yawei and Benini, Luca},
  journal={arXiv preprint arXiv:2509.11177},
  year={2025}
}

@article{liu2024eora,
  title={EoRA: Fine-tuning-free Compensation for Compressed LLM with Eigenspace Low-Rank Approximation},
  author={Liu, Shih-Yang and Khadkevich, Maksim and Fung, Nai Chit and Sakr, Charbel and Yang, Chao-Han Huck and Wang, Chien-Yi and Muralidharan, Saurav and Yin, Hongxu and Cheng, Kwang-Ting and Kautz, Jan and others},
  journal={arXiv preprint arXiv:2410.21271},
  year={2024}
}

@inproceedings{
sengupta2025you,
title={You Only Prune Once: Designing Calibration-Free Model Compression With Policy Learning},
author={Ayan Sengupta and Siddhant Chaudhary and Tanmoy Chakraborty},
booktitle={The Thirteenth International Conference on Learning Representations},
year={2025},
url={https://openreview.net/forum?id=5RZoYIT3u6}
}

@inproceedings{
li2025trthepruner,
title={T\'yr-the-Pruner: Structural Pruning {LLM}s via Global Sparsity Distribution Optimization},
author={Guanchen Li and Yixing Xu and Zeping Li and Ji Liu and Xuanwu Yin and Dong Li and Emad Barsoum},
booktitle={The Thirty-ninth Annual Conference on Neural Information Processing Systems},
year={2025},
url={https://openreview.net/forum?id=rAuRLePL2R}
}

@inproceedings{yang2025let,
  title={Let llm tell what to prune and how much to prune},
  author={Yang, Mingzhe and Lin, Sihao and Li, Changlin and Chang, Xiaojun},
  booktitle={Forty-second International Conference on Machine Learning},
  year={2025}
}

@article{guo2025slimllm,
  title={SlimLLM: Accurate structured pruning for large language models},
  author={Guo, Jialong and Chen, Xinghao and Tang, Yehui and Wang, Yunhe},
  journal={arXiv preprint arXiv:2505.22689},
  year={2025}
}

@inproceedings{men2025shortgpt,
  title={Shortgpt: Layers in large language models are more redundant than you expect},
  author={Men, Xin and Xu, Mingyu and Zhang, Qingyu and Yuan, Qianhao and Wang, Bingning and Lin, Hongyu and Lu, Yaojie and Han, Xianpei and Chen, Weipeng},
  booktitle={Findings of the Association for Computational Linguistics: ACL 2025},
  pages={20192--20204},
  year={2025}
}

@inproceedings{
hinostroza2026rethinking,
title={Rethinking Layer Relevance in Large Language Models Beyond Cosine Similarity},
author={Cristian Hinostroza and Rodrigo Toro Icarte and Christ Devia and Andres Carvallo De Ferari and Eugenio Herrera-Berg and Denis Parra and Jorge F Silva},
booktitle={The Fourteenth International Conference on Learning Representations},
year={2026},
url={https://openreview.net/forum?id=mRLnS8jQWt}
}

@inproceedings{
cai2025llamaflex,
title={{LL}aMaFlex: Many-in-one {LLM}s via Generalized Pruning and Weight Sharing},
author={Ruisi Cai and Saurav Muralidharan and Hongxu Yin and Zhangyang Wang and Jan Kautz and Pavlo Molchanov},
booktitle={The Thirteenth International Conference on Learning Representations},
year={2025},
url={https://openreview.net/forum?id=AyC4uxx2HW}
}

@inproceedings{
saukh2026cut,
title={Cut Less, Fold More: Model Compression through the Lens of Projection Geometry},
author={Olga Saukh and Dong Wang and Haris {\v{S}}iki{\'c} and Yun Cheng and Lothar Thiele},
booktitle={The Fourteenth International Conference on Learning Representations},
year={2026},
url={https://openreview.net/forum?id=JV9CEtKLQF}
}

@inproceedings{
le2025probe,
title={Probe Pruning: Accelerating {LLM}s through Dynamic Pruning via Model-Probing},
author={Qi Le and Enmao Diao and Ziyan Wang and Xinran Wang and Jie Ding and Li Yang and Ali Anwar},
booktitle={The Thirteenth International Conference on Learning Representations},
year={2025},
url={https://openreview.net/forum?id=WOt1owGfuN}
}

@article{yuan2023asvd,
  title={Asvd: Activation-aware singular value decomposition for compressing large language models},
  author={Yuan, Zhihang and Shang, Yuzhang and Song, Yue and Yang, Dawei and Wu, Qiang and Yan, Yan and Sun, Guangyu},
  journal={arXiv preprint arXiv:2312.05821},
  year={2023}
}

@inproceedings{
wang2025svdllm,
title={{SVD}-{LLM}: Truncation-aware Singular Value Decomposition for Large Language Model Compression},
author={Xin Wang and Yu Zheng and Zhongwei Wan and Mi Zhang},
booktitle={The Thirteenth International Conference on Learning Representations},
year={2025},
url={https://openreview.net/forum?id=LNYIUouhdt}
}

@inproceedings{
lin2025modegpt,
title={MoDe{GPT}: Modular Decomposition for Large Language Model Compression},
author={Chi-Heng Lin and Shangqian Gao and James Seale Smith and Abhishek Patel and Shikhar Tuli and Yilin Shen and Hongxia Jin and Yen-Chang Hsu},
booktitle={The Thirteenth International Conference on Learning Representations},
year={2025},
url={https://openreview.net/forum?id=8EfxjTCg2k}
}

@inproceedings{
chiang2026uniql,
title={Uni{QL}: Unified Quantization and Low-rank Compression for Adaptive Edge {LLM}s},
author={Hung-Yueh Chiang and Chi-Chih Chang and Yu-Chen Lu and Chien-Yu Lin and Kai-Chiang Wu and Mohamed S. Abdelfattah and Diana Marculescu},
booktitle={The Fourteenth International Conference on Learning Representations},
year={2026},
url={https://openreview.net/forum?id=iOGu4wtDTF}
}

@inproceedings{
hu2026saessvd,
title={{SAES}-{SVD}: Self-Adaptive Suppression of Accumulated and Local Errors for {SVD}-based {LLM} Compression},
author={Xing Hu and Zukang Xu and Zhixuan Chen and Yuan Cheng and Dawei Yang},
booktitle={The Fourteenth International Conference on Learning Representations},
year={2026},
url={https://openreview.net/forum?id=KMAYsQO8pU}
}

@article{bai2025ressvd,
  title={Ressvd: Residual compensated svd for large language model compression},
  author={Bai, Haolei and Jian, Siyong and Liang, Tuo and Yin, Yu and Wang, Huan},
  journal={arXiv preprint arXiv:2505.20112},
  year={2025}
}

@inproceedings{
mozaffari2025slim,
title={{SL}iM: One-shot Quantization and Sparsity with Low-rank Approximation for {LLM} Weight Compression},
author={Mohammad Mozaffari and Amir Yazdanbakhsh and Maryam Mehri Dehnavi},
booktitle={Forty-second International Conference on Machine Learning},
year={2025},
url={https://openreview.net/forum?id=4UfRP8MopP}
}

@article{koike2025latentllm,
  title={Latentllm: Attention-aware joint tensor compression},
  author={Koike-Akino, Toshiaki and Chen, Xiangyu and Liu, Jing and Wang, Ye and Brand, Matthew and others},
  journal={arXiv preprint arXiv:2505.18413},
  year={2025}
}

@article{li2025adasvd,
  title={Adasvd: Adaptive singular value decomposition for large language models},
  author={Li, Zhiteng and Xia, Mingyuan and Zhang, Jingyuan and Hui, Zheng and Qin, Haotong and Kong, Linghe and Zhang, Yulun and Yang, Xiaokang},
  journal={arXiv preprint arXiv:2502.01403},
  year={2025}
}

@inproceedings{
wang2025basis,
title={Basis Sharing: Cross-Layer Parameter Sharing for Large Language Model Compression},
author={Jingcun Wang and Yu-Guang Chen and Ing-Chao Lin and Bing Li and Grace Li Zhang},
booktitle={The Thirteenth International Conference on Learning Representations},
year={2025},
url={https://openreview.net/forum?id=gp32jvUquq}
}

@inproceedings{
li2025moesvd,
title={MoE-{SVD}: Structured Mixture-of-Experts {LLM}s Compression via Singular Value Decomposition},
author={Wei Li and Lujun Li and Hao Gu and You-Liang Huang and Mark G. Lee and Shengjie Sun and Wei Xue and Yike Guo},
booktitle={Forty-second International Conference on Machine Learning},
year={2025},
url={https://openreview.net/forum?id=acJ3vdFljk}
}

@inproceedings{huang2025sola,
  title={Sola: Leveraging soft activation sparsity and low-rank decomposition for large language model compression},
  author={Huang, Xinhao and Huang, You-Liang and Wen, Zeyi},
  booktitle={Proceedings of the AAAI Conference on Artificial Intelligence},
  volume={39},
  number={16},
  pages={17494--17502},
  year={2025}
}

@inproceedings{
wang2025qsvd,
title={{QSVD}: Efficient Low-rank Approximation for Unified Query-Key-Value Weight Compression in Low-Precision Vision-Language Models},
author={Yutong Wang and Haiyu Wang and Sai Qian Zhang},
booktitle={The Thirty-ninth Annual Conference on Neural Information Processing Systems},
year={2025},
url={https://openreview.net/forum?id=sEFDhxF1mG}
}

@inproceedings{
li2025svdquant,
title={{SVDQ}uant: Absorbing Outliers by Low-Rank Component for 4-Bit Diffusion Models},
author={Muyang Li and Yujun Lin and Zhekai Zhang and Tianle Cai and Junxian Guo and Xiuyu Li and Enze Xie and Chenlin Meng and Jun-Yan Zhu and Song Han},
booktitle={The Thirteenth International Conference on Learning Representations},
year={2025},
url={https://openreview.net/forum?id=vWR3KuiQur}
}

@inproceedings{
zhang2026qronos,
title={Qronos: Correcting the Past by Shaping the Future... in Post-Training Quantization},
author={Shihao Zhang and Haoyu Zhang and Ian Colbert and Rayan Saab},
booktitle={The Fourteenth International Conference on Learning Representations},
year={2026},
url={https://openreview.net/forum?id=7axclBCYul}
}

@inproceedings{
frantar2023optq,
title={{OPTQ}: Accurate Quantization for Generative Pre-trained Transformers},
author={Elias Frantar and Saleh Ashkboos and Torsten Hoefler and Dan Alistarh},
booktitle={The Eleventh International Conference on Learning Representations },
year={2023},
url={https://openreview.net/forum?id=tcbBPnfwxS}
}

@inproceedings{
li2025gptaq,
title={{GPTAQ}: Efficient Finetuning-Free Quantization for Asymmetric Calibration},
author={Yuhang Li and Ruokai Yin and Donghyun Lee and Shiting Xiao and Priyadarshini Panda},
booktitle={Forty-second International Conference on Machine Learning},
year={2025},
url={https://openreview.net/forum?id=QdELyl0FST}
}

@article{chee2023quip,
  title={Quip: 2-bit quantization of large language models with guarantees},
  author={Chee, Jerry and Cai, Yaohui and Kuleshov, Volodymyr and De Sa, Christopher M},
  journal={Advances in neural information processing systems},
  volume={36},
  pages={4396--4429},
  year={2023}
}

@inproceedings{
liu2025spinquant,
title={SpinQuant: {LLM} Quantization with Learned Rotations},
author={Zechun Liu and Changsheng Zhao and Igor Fedorov and Bilge Soran and Dhruv Choudhary and Raghuraman Krishnamoorthi and Vikas Chandra and Yuandong Tian and Tijmen Blankevoort},
booktitle={The Thirteenth International Conference on Learning Representations},
year={2025},
url={https://openreview.net/forum?id=ogO6DGE6FZ}
}

@inproceedings{
liang2026paroquant,
title={ParoQuant: Pairwise Rotation Quantization for Efficient Reasoning {LLM} Inference},
author={Yesheng Liang and Haisheng Chen and Song Han and Zhijian Liu},
booktitle={The Fourteenth International Conference on Learning Representations},
year={2026},
url={https://openreview.net/forum?id=1USeVjsKau}
}

@inproceedings{
arai2025quantization,
title={Quantization Error Propagation: Revisiting Layer-Wise Post-Training Quantization},
author={Yamato Arai and Yuma Ichikawa},
booktitle={The Thirty-ninth Annual Conference on Neural Information Processing Systems},
year={2025},
url={https://openreview.net/forum?id=a3l3K9khbL}
}

@inproceedings{
ji2025beware,
title={Beware of Calibration Data for Pruning Large Language Models},
author={Yixin Ji and Yang Xiang and Juntao Li and Qingrong Xia and Ping Li and Xinyu Duan and Zhefeng Wang and Min Zhang},
booktitle={The Thirteenth International Conference on Learning Representations},
year={2025},
url={https://openreview.net/forum?id=x83w6yGIWb}
}

@inproceedings{
he2025preserving,
title={Preserving {LLM} Capabilities through Calibration Data Curation: From Analysis to Optimization},
author={Bowei He and Lihao Yin and Huiling Zhen and Shuqi LIU and Han Wu and Xiaokun Zhang and Mingxuan Yuan and Chen Ma},
booktitle={The Thirty-ninth Annual Conference on Neural Information Processing Systems},
year={2025},
url={https://openreview.net/forum?id=Tf9eoTIIjh}
}

@inproceedings{williams-aletras-2024-impact,
    title = "On the Impact of Calibration Data in Post-training Quantization and Pruning",
    author = "Williams, Miles  and
      Aletras, Nikolaos",
    editor = "Ku, Lun-Wei  and
      Martins, Andre  and
      Srikumar, Vivek",
    booktitle = "Proceedings of the 62nd Annual Meeting of the Association for Computational Linguistics (Volume 1: Long Papers)",
    month = aug,
    year = "2024",
    address = "Bangkok, Thailand",
    publisher = "Association for Computational Linguistics",
    url = "https://aclanthology.org/2024.acl-long.544/",
    doi = "10.18653/v1/2024.acl-long.544",
    pages = "10100--10118"
}
\bibliographystyle{colm2026_conference}

\newpage

\appendix

\section{LLM Usage}

We use LLM coding assistants in the implementation of our methods and data analysis scripts. Minor usage for Tables and Latex formatting. Everything else from idea generation, problem formulation and manuscript writing has been conducted mainly by the authors.

\section{Detailed Experimental Settings}\label{appdx:exp}

\textbf{Experimental Setup.}
We follow the same experimental setting as UniQL~\citep{chiang2026uniql} as we build on their framework. By default, we reuse the same seed, sequence length of 2048, and perform our evaluations in BF16. We also implement MoDeGPT based on their paper details~\citep{lin2025modegpt} and use their default experimental configuration setting. We run our experiments on two NVIDIA RTX Ada 6000 GPUs with 48 GB of memory each. 

\textbf{Evaluation Benchmarks and Criteria.}
We follow the UniQL convention and conduct our main evaluations on five 0-shot down stream tasks from LM-Eval Harness~\citep{eval-harness}:  PiQA, Winogrande, Arc-Easy, Arc-Challenge, Hellaswag. We report the accuracy for the first three benchmarks, and normalized accuracy for Arc-Challenge and Hellaswag. We also report the 5-shot MMLU results. Our evaluations are conducted using a single GPU with fixed settings following the standard configuration of lm-eval-harness~\citep{eval-harness}. Our method does not involve any backpropagation since it targets training-free frameworks. We report single run results following the convention of our baselines SVD-LLM, MoDeGPT, UniQL. In a separate experiment, we ablate the efficacy of our method through repeated runs of the experiment using different calibration data sampling seeds to validate the outcomes.

\textbf{MoDeGPT Implementation.}
Due to a lack of open-source implementation, we implement MoDeGPT following their paper details. Llama-2-7B is the common model between their work and ours, and thus we use it for validating our implementation and cross-reference the results. We reproduce the average LM-Eval numbers when the Llama-2-7B is compressed at 20\%,30\%, and 40\% from their Table 11. The averages from our implementation compared to MoDeGPT were 65.4\%(65.62\%); 62.4\%(62.02\%); 56.9\%(57.58\%), with a maximum deviation of 0.7\%. We use their default setting of 128 Wiki Calibration data samples.

\subsection{Metrics}

\textbf{Normalized Mean Squared Error (NMSE).} To evaluate the residual calibration error, we measure the representation error between activations, given by the Normalized Mean Square Error (NMSE):
\begin{equation*}
\text{NMSE}(x, \hat{x}) = \frac{\sum (x-\hat{x})^2}{\sum x^2},
\end{equation*}
where $x$ and $\hat{x}$ denote the activations of the original and compressed models respectively. NMSE provides a scale-invariant measure of reconstruction error, making it suitable for comparing errors across layers of varying sizes.

\section{Ablation Studies and Further Experimental Analysis} \label{appdx:exps}

\subsection{Residual Calibration Errors Analysis} \label{appdx:calib}

\begin{figure}[!ht]
\begin{center}
{\includegraphics[,width = 0.98\textwidth]{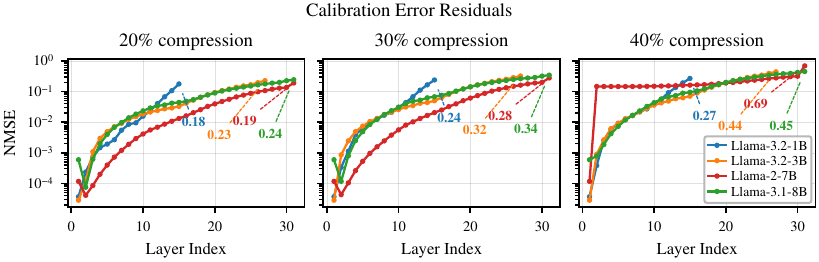}}
\end{center}
\vspace{-2ex}
\caption{Residual Calibration Errors in Llama models measured through per-layer NMSE}
\label{fig:calib_llamas}
\end{figure}

\begin{figure}[!ht]
\begin{center}
{\includegraphics[,width = 0.98\textwidth]{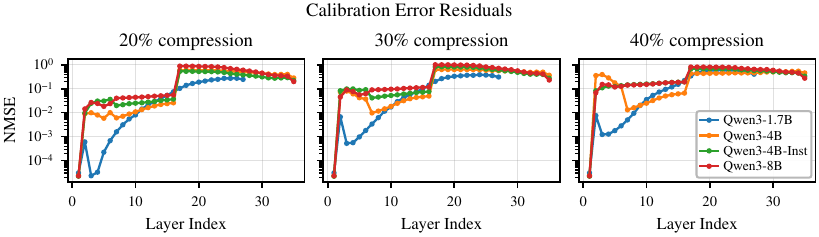}}
\end{center}
\vspace{-2ex}
\caption{Residual Calibration Errors in Qwen3 models measured through per-layer NMSE}
\label{fig:calib_qwen3}
\end{figure}

 Figures \ref{fig:calib_llamas} and \ref{fig:calib_qwen3} showcase the residual calibration errors\footnote{By residual, we mean the drift present in the activations at each layer, not a cumulative sum of per-layer errors.} measured through the per-layer NMSE for the Llama and Qwen3 models, respectively. We observe that the Qwen3 models exhibit a more variable behavior compared to the Llama models, especially in the shallower layers. This was also reflected when we modeled the layer importance of Qwen3 models in Figure~\ref{fig:div_error_2}, and found that the first layer exhibits significantly larger BI Score compared to succeeding layers. 

\begin{table}[t!]
\centering
\small
\setlength{\tabcolsep}{7pt}
\renewcommand{\arraystretch}{1.12}
\begin{tabular}{@{}lcccc@{}}
\toprule
& \multicolumn{2}{c}{\textbf{15\% Compression}}
& \multicolumn{2}{c}{\textbf{40\% Compression}} \\
\cmidrule(lr){2-3}
\cmidrule(l){4-5}
\textbf{Model}
& \textbf{First Layer}
& \textbf{Last Layer}
& \textbf{First Layer}
& \textbf{Last Layer} \\
\midrule
Qwen3 1.7B
& 0.9999
& 0.9926
& 0.9990
& 0.9802 \\

Qwen3 4B
& 0.9999
& 0.9318
& 0.9998
& 0.8181 \\

Qwen3 8B
& 0.9995
& 0.8928
& 0.9988
& 0.7140 \\
\bottomrule
\end{tabular}
\caption{
First- and last-layer BI scores for Qwen3 models at 15\% and 40\% compression.
}
\label{tab:qwen_first_last}
\end{table}

\subsection{Layer Ratio Misalignment Error Analysis} \label{appdx:qwen_bi}

\begin{figure}[!t]
    \centering
    \begin{subfigure}[b]{0.32\textwidth}
        \centering
        \includegraphics[width=\textwidth]{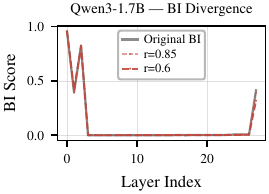} 
        \label{fig:div_17b}
    \end{subfigure}
    \begin{subfigure}[b]{0.32\textwidth}
        \centering
        \includegraphics[width=\textwidth]{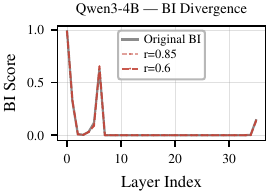} 
        \label{fig:div_4b}
    \end{subfigure}
    \begin{subfigure}[b]{0.32\textwidth}
        \centering
        \includegraphics[width=\textwidth]{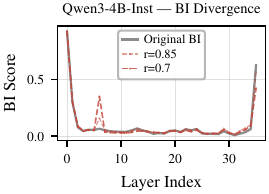}
        \label{fig:div_06b}
    \end{subfigure}
    
    \caption{BI Score Divergence pre and post compression for Qwen3-\{1.7B, 4B, 4B-Inst-2507\}}
    \label{fig:div_error_2}
\end{figure}

\begin{figure}[!t]
    \centering
    \begin{subfigure}[b]{0.32\textwidth}
        \centering
        \includegraphics[width=\textwidth]{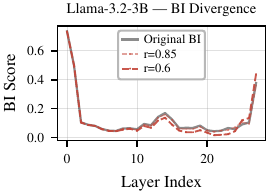}
        \label{fig:div_3b}
    \end{subfigure}
    \begin{subfigure}[b]{0.32\textwidth}
        \centering
        \includegraphics[width=\textwidth]{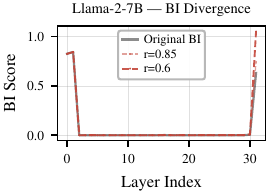} 
        \label{fig:div_7b}
    \end{subfigure}
    \begin{subfigure}[b]{0.32\textwidth}
        \centering
        \includegraphics[width=\textwidth]{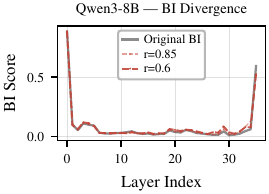}
        \label{fig:div_q8b}
    \end{subfigure}
    \caption{BI Score Divergence pre and post compression for Llama-\{3.2-3b, 2-7B\}, Qwen3-8B}
    \label{fig:div_error_3}
    
\end{figure}

Figures \ref{fig:div_error_2} and \ref{fig:div_error_3} illustrate rank ratio allocation divergence analysis across Llama and Qwen3 models before and after compression. We observe that the Qwen3 models are generally more stable and their initial BI estimates tend to hold to a large extent post-calibration. This can be attributed to the disproportionate importance assigned to the first and last layers of the Qwen3 models.  
Table \ref{tab:qwen_first_last} shows the Block Influence (BI) scores assigned to the first and last layers of the different Qwen3 models, demonstrating how their dominant BI scores lead them to be assigned high retention ratios (close to 100\% retention). Qwen3's large vocabulary size of 151k can be a reason for the disproportionate first layer importance.

\subsection{Further Detailed Analysis for Main Experiments.}

\begin{figure}[!t]
\begin{center}
{\includegraphics[,width = 0.95\textwidth]{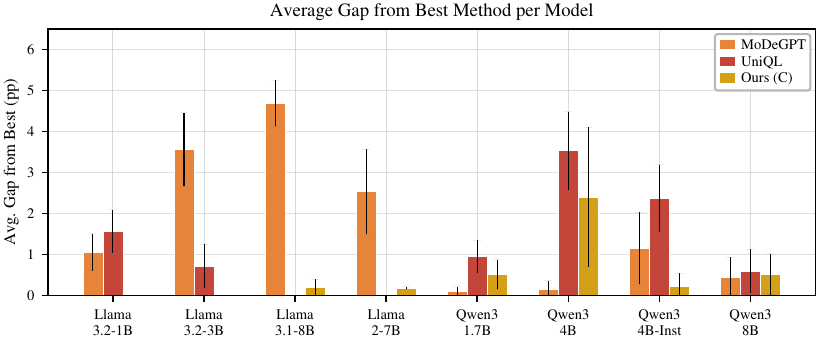}}
\end{center}
\vspace{-2ex}
\caption{Average Gap from Best Analysis aggregated for every model (Lower is Better)}
\label{fig:gap_by_model}
\end{figure}

\textbf{Average Gap from best. }In Figure \ref{fig:gap_by_model}, we show the average gap for each method in terms of percentage point from the best 0-shot LM-Eval average incurred by any method \textbf{on a per-model basis}, where lower is better. The plot demonstrates that our calibration correction is performant on the Llama models, yet far from the MoDeGPT method on the smallest Qwen3 models. The Qwen3 model shows competitive results across all three methods. 

\textbf{Breakdown of 0-shot LM-Eval Average Results.} In Tables \ref{tab:0_shot_085} and \ref{tab:0_shot_07}, we break down the LM-Eval 0-shot average performance reported in the main Table~\ref{tab:zero-shot} by benchmark for Llama-\{3.2-1B, 3.1-8B\} and Qwen3-8B at 15\% and 30\% compression ratios, respectively.

\textbf{MMLU Results.} We also report 5-shot MMLU evaluation for our method against the baseline UniQL~\citep{chiang2026uniql} for the best options identified in Table \ref{tab:zero-shot} at 15\% and 30\% compression rates. The results indicate that the MMLU evaluations follow the same pattern as that of the LM-Eval Average.

\textbf{NMSE Reduction.} In Table \ref{tab:nmse}, we report the drop in NMSE after applying the calibration correction across the 3 last layers for the Llama 3.1-8B and Qwen3-4B at 25\% compression.

\begin{table}[t]
\small
\centering
\begin{tabular}{ll|ccccc|c}
\toprule
\textbf{Model} & \textbf{Method} & \textbf{PiQA} & \textbf{WinoGr} & \textbf{Arc-e} & \textbf{Arc-c} & \textbf{Hswag} & \textbf{Avg} \\
\midrule\midrule
\multirow{4}{*}{Llama 3.2-1B}
  & MoDeGPT   & 69.42& 60.54& 57.20& 31.14& 54.11& 54.48\\
  & UniQL     & 69.86& 58.72& 52.99& 33.02& 53.38& 53.59\\
  \cline{2-8}
  & \cellcolor{second}Ours (C)  & \cellcolor{second}70.57& \cellcolor{second}60.54& \cellcolor{second}56.65& \cellcolor{second}32.85& \cellcolor{second}54.70& \cellcolor{second}55.06\\
  & \cellcolor{first}Ours (C+R)& \cellcolor{first}71.82& \cellcolor{first}59.67& \cellcolor{first}57.62& \cellcolor{first}32.00& \cellcolor{first}55.28& \cellcolor{first}\textbf{55.28}\\
\midrule\midrule
\multirow{4}{*}{Llama 3.1-8B}
  & MoDeGPT   & 75.19& 71.51& 72.43& 42.92& 71.50& 66.71\\
  & UniQL     & 76.17& 73.24& 77.53& 50.77& 75.28& \textbf{70.60}\\
  \cline{2-8}
  & \cellcolor{second}Ours (C)  &\cellcolor{second} 77.48& \cellcolor{second}71.03&\cellcolor{second} 78.87&\cellcolor{second} 51.62&\cellcolor{second} 73.68&\cellcolor{second} 70.54\\
  & \cellcolor{first}Ours (C+R)& \cellcolor{first}77.15& \cellcolor{first}71.51& \cellcolor{first}78.58& \cellcolor{first}51.96& \cellcolor{first}73.62& \cellcolor{first}70.56\\
\midrule\midrule
\multirow{4}{*}{Qwen 3-8B}
  & MoDeGPT   & 73.67& 66.46& 75.80& 50.34& 70.24& \textbf{67.30}\\
  & UniQL     & 73.94& 67.09& 70.71& 47.01& 71.29& 66.01\\
  \cline{2-8}
  & \cellcolor{second}Ours (C)  & \cellcolor{second}73.61& \cellcolor{second}65.75& \cellcolor{second}72.01&\cellcolor{second} 49.23 & \cellcolor{second} 69.97& \cellcolor{second}66.11\\
  & \cellcolor{first}Ours (C+R)& \cellcolor{first}73.67& \cellcolor{first}66.54& \cellcolor{first}72.18& \cellcolor{first}49.23& \cellcolor{first}70.04&\cellcolor{first}66.33 \\
\midrule\bottomrule
\end{tabular}
\caption{Breakdown of the LM-Eval Average Results from Table \ref{tab:zero-shot} at 15\% Compression. }
\label{tab:0_shot_085}
\end{table}
\begin{table}[t]
\small
\centering
\begin{tabular}{ll|ccccc|c}
\toprule
\textbf{Model} & \textbf{Method} & \textbf{PiQA} & \textbf{WinoGr} & \textbf{Arc-e} & \textbf{Arc-c} & \textbf{Hswag} & \textbf{Avg} \\
\midrule\midrule
\multirow{4}{*}{Llama 3.2-1B}
  & MoDeGPT   & 60.12 & 56.59& 39.73& 24.66& 38.36& 43.89\\
  & UniQL     & 59.68& 53.83& 34.81& 30.80& 37.83& 43.30\\
  \cline{2-8}
  & \cellcolor{second}Ours (C)  & \cellcolor{second}63.38& \cellcolor{second}53.75& \cellcolor{second}43.35& \cellcolor{second}27.65& \cellcolor{second}39.61& \cellcolor{second}45.55\\
  & \cellcolor{first}Ours (C+R)& \cellcolor{first}63.55& \cellcolor{first}54.38& \cellcolor{first}44.28& \cellcolor{first}28.58& \cellcolor{first}40.58& \cellcolor{first}\textbf{46.27}\\
\midrule\midrule
\multirow{4}{*}{Llama 3.1-8B}
  & MoDeGPT   & 68.39& 68.11& 61.20& 34.30& 58.75& 58.15\\
  & UniQL     & 71.82& 70.88& 66.62& 41.81& 64.90& 63.21\\
  \cline{2-8}
  & \cellcolor{second}Ours (C)  & \cellcolor{second}72.31& \cellcolor{second}66.46& \cellcolor{second}69.91& \cellcolor{second}42.06& \cellcolor{second}62.88& \cellcolor{second}62.72\\
  & \cellcolor{first}Ours (C+R)& \cellcolor{first}72.74& \cellcolor{first}66.77& \cellcolor{first}70.24& \cellcolor{first}44.88& \cellcolor{first}63.68& \cellcolor{first}\textbf{63.66}\\
\midrule\midrule
\multirow{4}{*}{Qwen 3-8B}
  & MoDeGPT   & 65.89& 61.25& 59.26& 34.30& 54.56& 55.05\\
  & UniQL     & 67.36& 60.06& 55.98& 36.77& 58.12& 55.66\\
  \cline{2-8}
  & \cellcolor{second} Ours (C)  & \cellcolor{second}67.90& \cellcolor{second}59.98& \cellcolor{second}58.96& \cellcolor{second}38.48& \cellcolor{second}55.52& \cellcolor{second}56.17\\
  & \cellcolor{first}Ours (C+R)& \cellcolor{first}68.28& \cellcolor{first}62.75& \cellcolor{first}57.79& \cellcolor{first}38.23& \cellcolor{first}55.76& \cellcolor{first}\textbf{56.56}\\
\midrule\bottomrule
\end{tabular}
\caption{Breakdown of the LM-Eval Average Results from Table \ref{tab:zero-shot} at 30\% Compression. }
\label{tab:0_shot_07}
\end{table}
\begin{table}[t]
\centering
\resizebox{0.9\columnwidth}{!}{%
\begin{tabular}{cl cccc cccc}
\toprule
\multirow{2}{*}{\shortstack{Compression\\Ratio}} & \multirow{2}{*}{Method} 
  & \multicolumn{4}{c}{LLaMA} 
  & \multicolumn{4}{c}{Qwen 3} \\
\cmidrule(lr){3-6} \cmidrule(lr){7-10}
& & 3.2-1B & 3.2-3B & 3.1-8B & 2-7B & 8B \\
\midrule\midrule
-- & Original (Uncompressed) & 31.17 & 56.50 & 65.35 & 45.86 & 74.90  \\
\midrule\midrule
\multirow{4}{*}{15\%} 
  & UniQL~\citep{chiang2026uniql} & 23.52 & 50.88 & \textbf{59.93} & 40.73 & 61.83 \\
  \cline{2-10}
  & \cellcolor{second}Ours (C)     &  \cellcolor{second}23.93 & \cellcolor{second}50.69 & \cellcolor{second}59.91 & \cellcolor{second}\textbf{43.57} & \cellcolor{second}65.80 \\
  & \cellcolor{first}Ours (C+R)     &  \cellcolor{first}\textbf{25.3} & \cellcolor{first} \textbf{52.24} & \cellcolor{first} 59.46 &\cellcolor{first} 43.40 & \cellcolor{first}\textbf{65.88} \\
\midrule\midrule
\multirow{4}{*}{30\%} 
  & UniQL~\citep{chiang2026uniql}   & 23.1 & 38.17 & \textbf{45.58} & 35.89 & 32.91  \\
  \cline{2-10}
  & \cellcolor{second}Ours (C)    & \cellcolor{second}\textbf{25.53} &  \cellcolor{second}38.78 & \cellcolor{second}41.18 & \cellcolor{second}\textbf{49.05} & \cellcolor{second}36.48 \\
  & \cellcolor{first}Ours (C+R)     &  \cellcolor{first}23.79 &  \cellcolor{first}\textbf{39.85} & \cellcolor{first} 41.21 & \cellcolor{first}46.34 & \cellcolor{first}\textbf{36.83} \\
\bottomrule
\end{tabular}%
}
\caption{MMLU 5-shot score accuracy comparing our approach against the baseline UniQL~\citep{chiang2026uniql}. (C) implies calibration correction; (R) implies layer rank correction. The (C+R) scores are for the best LM-Eval configuration from Table \ref{tab:zero-shot}.}
\label{tab:5-shot}
\end{table}

\begin{table*}[t]
\centering
\small
\setlength{\tabcolsep}{7pt}
\renewcommand{\arraystretch}{1.12}
\begin{tabular}{@{}lccccc@{}}
\toprule
\textbf{Model}
& \textbf{Layer} $\mathbf{L}-2$
& \textbf{Layer} $\mathbf{L}-1$
& \textbf{Layer} $\mathbf{L}$
& \textbf{Overall Avg.}
& \textbf{Cumulative} \\
\midrule

Llama 3.1-8B
& 33.72 / \textbf{32.41}
& 25.26 / \textbf{24.18}
& 0.74 / \textbf{0.58}
& 14.49 / \textbf{14.46}
& 463.67 / \textbf{462.73} \\

Qwen3-4B
& 23.87 / \textbf{23.00}
& 34.14 / \textbf{34.11}
& 11.22 / \textbf{7.35}
& 9.30 / \textbf{8.68}
& 334.83 / \textbf{312.36} \\

\bottomrule
\end{tabular}
\caption{
Layer-wise and aggregate NMSE at 25\% compression.
Entries report\textbf{ UniQL  / Ours (C)}, with the lower value in bold.
All values are scaled by $10^{-3}$, and $\mathbf{L}$ denotes the last layer.
}
\label{tab:nmse}
\end{table*}

\subsection{Multi-Stage BI Score Ablation}
\begin{figure}[p]
    \centering
    \includegraphics[width=0.32\columnwidth]{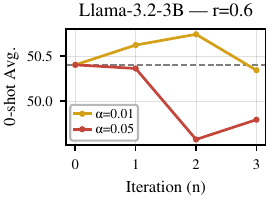}
    \includegraphics[width=0.32\columnwidth]{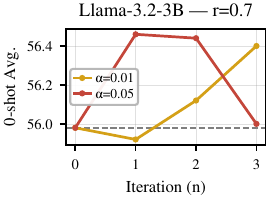}
    \includegraphics[width=0.32\columnwidth]{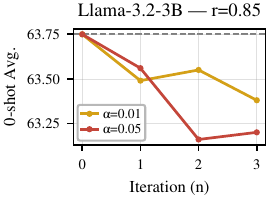}
    
    \caption{Avg. 0-shot scores for Llama-3.2-3B post BI-scores correction (Higher is better).}
    \label{fig:3b_bi_iter}
\end{figure}

\begin{figure}[p]
    \centering
    \includegraphics[width=0.32\columnwidth]{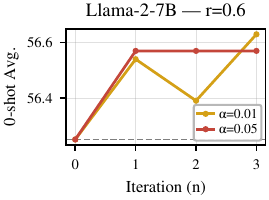}
    \includegraphics[width=0.32\columnwidth]{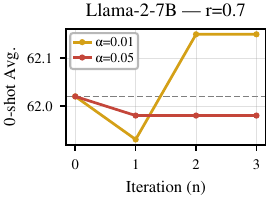}
    \includegraphics[width=0.32\columnwidth]{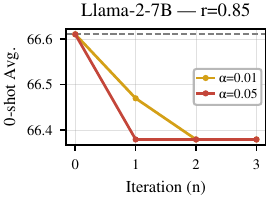}
    
    \caption{Avg. 0-shot scores for Llama-2-7B post BI-scores correction (Higher is better).}
    \label{fig:7b_bi_iter}
\end{figure}

\begin{figure}[p]
    \centering
    \includegraphics[width=0.32\columnwidth]{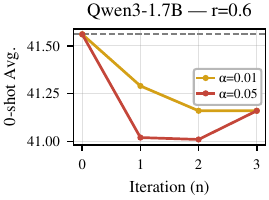}
    \includegraphics[width=0.32\columnwidth]{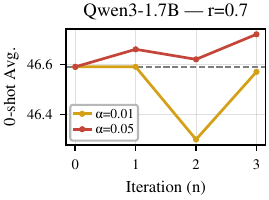}
    \includegraphics[width=0.32\columnwidth]{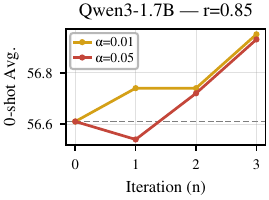}
    
    \caption{Avg. 0-shot scores for Qwen3-1.7B post BI-scores correction (Higher is better).}
    \label{fig:q1.7b_bi_iter}
\end{figure}

\begin{figure}[p]
    \centering
    \includegraphics[width=0.32\columnwidth]{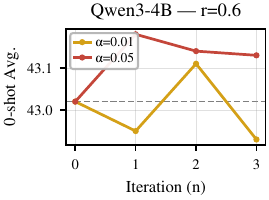}
    \includegraphics[width=0.32\columnwidth]{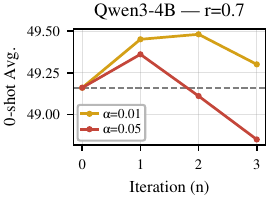}
    \includegraphics[width=0.32\columnwidth]{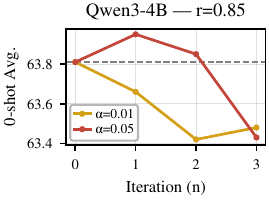}
    
    \caption{Avg. 0-shot scores for Qwen3-4B post BI-scores correction (Higher is better).}
    \label{fig:q4b_bi_iter}
\end{figure}

\begin{figure}[p]
    \centering
    \includegraphics[width=0.32\columnwidth]{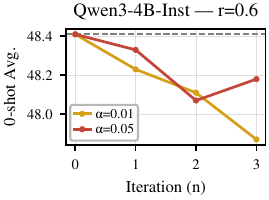}
    \includegraphics[width=0.32\columnwidth]{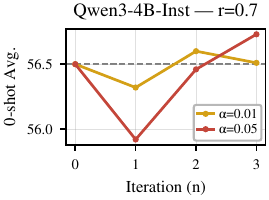}
    \includegraphics[width=0.32\columnwidth]{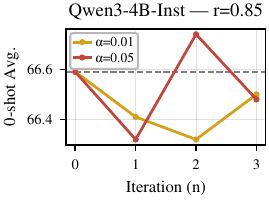}
    
    \caption{Avg. 0-shot scores for Qwen3-4B-Instruct-2507 post BI-scores correction (Higher is better).}
    \label{fig:q4bInst_bi_iter}
\end{figure}

\begin{figure}[!t]
    \centering
    \includegraphics[width=0.32\columnwidth]{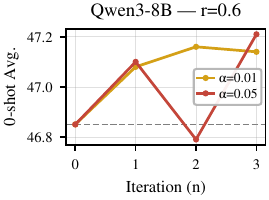}
    \includegraphics[width=0.32\columnwidth]{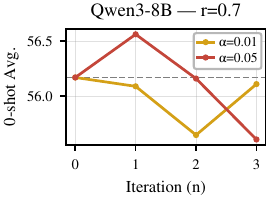}
    \includegraphics[width=0.32\columnwidth]{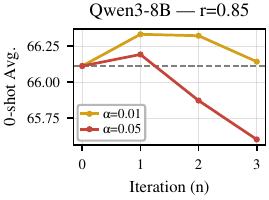}
    
    \caption{Avg. 0-shot Scores for Qwen3-8B post BI-scores correction (Higher is better).}
    \label{fig:q8b_bi_iter}
\end{figure}

Figures \ref{fig:3b_bi_iter}-\ref{fig:q8b_bi_iter} showcase the change in 0-shot LM-Eval Average score induced by each model when it undergoes up to three rounds of iterative compression and rank ratio refinement at $\alpha\in\{0.05, 0.01\}$ and rank ratios $r\in\{0.6, 0.7, 0.85\}$. The plots showcase that a small number of iterations is sufficient to align the layer ratios with the compressed model representations to realize performance gains.

\clearpage

\subsection{Hyperparameter Ablation for $N$ and $\alpha$ }

We perform a grid search to ablate the different hyperparameter effects of $\alpha$ and $N$ aggregated across different model categories across 3 compression ratios 15\%, 30\%, and 40\%. 

\textbf{$\mathbf{\textit{N}}$ Effect. }In Table \ref{tab:N_effect}, we first ablate the effect of $N$ through recording the mean 0-shot average improvement when applying iterative refinement on top of corrective calibration (C). We run up to 3 iterations ($N$=3), and report the mean improvement over (C) across all configurations at the first and best $N$. We observe that a single iteration ($N$=1) is sufficient to attain up to 50\% improvement over (C).

\begin{table*}[t]
\centering
\small
\setlength{\tabcolsep}{12pt}
\renewcommand{\arraystretch}{1.15}
\begin{tabular}{@{}lccc@{}}
\toprule
\textbf{Setting}
& \shortstack{\textbf{Llama-3.2-1B /}\\\textbf{Llama-3.2-3B}}
& \shortstack{\textbf{Qwen3-4B /}\\\textbf{Qwen3-4B-Instruct}}
& \shortstack{\textbf{Llama-3.1-8B /}\\\textbf{Qwen3-8B}} \\
\midrule
Best $N$
& +0.29
& +0.11
& +0.39 \\
$N=1$
& +0.17
& +0.03
& +0.17 \\
\bottomrule
\end{tabular}
\caption{
Mean LM-Avg improvement of Ours (C+R) over Ours (C) across model groups.
Best $N$ selects the best-performing value from any $N$ for each model grouping.
}
\label{tab:N_effect}
\end{table*}

\textbf{$\alpha$ Effect. }In Table \ref{tab:alpha_effect}, we ablate the effects of varying $\alpha \in \{0.01, 0.05\}$ given a single iteration ($N$=1), and report the mean improvement across each collection of models. The results show that the average gains vary both within and across each model collection.

\begin{table*}[t]
\centering
\small
\setlength{\tabcolsep}{12pt}
\renewcommand{\arraystretch}{1.15}
\begin{tabular}{@{}lccc@{}}
\toprule
\textbf{Setting}
& \shortstack{\textbf{Llama-3.2-1B /}\\\textbf{Llama-3.2-3B}}
& \shortstack{\textbf{Qwen3-4B /}\\\textbf{Qwen3-4B-Instruct}}
& \shortstack{\textbf{Llama-3.1-8B /}\\\textbf{Qwen3-8B}} \\
\midrule
$\alpha=0.01$
& $+0.01$
& $-0.08$
& $+0.03$ \\

$\alpha=0.05$
& $-0.20$
& $-0.07$
& $+0.07$ \\

\midrule
Average
& $-0.10$
& $-0.07$
& $+0.05$ \\
\bottomrule
\end{tabular}
\caption{
Mean LM-Avg change across model groups at $N=1$.
}
\label{tab:alpha_effect}
\end{table*}

From this analysis, the takeaway is that a grid search over $\alpha$ can work best, and that a single refinement ($N$=1) can be sufficient to start realizing performance gains.

\subsection{Iterative Refinement (R) Ablation Summary }

We demonstrate the effects of (R) independent of (C) through the following:
\begin{itemize}
    \item The scaling experiment for the Qwen2.5-32B in Section~\ref{subsec:ablation} demonstrated that (R) alone can improve performance independent of (C). 
    \item The hyperparameter ablation demonstrated the contribution of iterative refinement across different model sizes where the average 0-shot accuracy improved by up to +0.39 pp compared to pure (C).
    \item In an experiment on Llama-3.2-1B, we applied one iteration of (R) without (C) to the baseline at $\alpha$ = 0.01, improving the average 0-shot accuracy from 41.27 to 41.41.
\end{itemize}

\begin{table*}[t!]
\centering
\small
\setlength{\tabcolsep}{5pt}
\renewcommand{\arraystretch}{1.12}
\begin{tabular}{@{}clccccccc@{}}
\toprule
\textbf{Comp.}
& \textbf{Method}
& \textbf{L-3.2-1B}
& \textbf{L-3.2-3B}
& \textbf{L-3.1-8B}
& \textbf{L-2-7B}
& \textbf{Q3-1.7B}
& \textbf{Q3-4B}
& \textbf{Q3-8B} \\
\midrule

\multirow{2}{*}{15\%}
& MoDeGPT
& 13.72
& \textbf{9.81}
& \textbf{7.80}
& 6.25
& \textbf{24.56}
& 15.95
& \textbf{11.42} \\
& Ours (C)
& \textbf{13.59}
& 9.83
& \textbf{7.80}
& \textbf{6.24}
& 26.92
& \textbf{15.43}
& 11.53 \\

\midrule

\multirow{2}{*}{30\%}
& MoDeGPT
& 32.54
& 14.27
& 10.79
& 7.59
& \textbf{28.33}
& 22.54
& 15.13 \\
& Ours (C)
& \textbf{32.29}
& \textbf{14.16}
& \textbf{10.77}
& \textbf{7.56}
& 29.86
& \textbf{20.44}
& \textbf{14.84} \\

\midrule

\multirow{2}{*}{40\%}
& MoDeGPT
& 44.21
& 22.32
& \textbf{15.19}
& 9.62
& 55.24
& 38.49
& 28.55 \\
& Ours (C)
& \textbf{41.70}
& \textbf{21.80}
& \textbf{15.19}
& \textbf{9.54}
& \textbf{54.07}
& \textbf{32.29}
& \textbf{26.56} \\
\bottomrule
\end{tabular}
\caption{
WikiText-2 perplexity across models and compression ratios (lower is better).
Bold indicates the better result. (L) and (Q3) denote Llama and Qwen3, respectively.
}
\label{tab:wiki}
\end{table*}

\begin{table*}[t!]
\centering
\small
\setlength{\tabcolsep}{5pt}
\renewcommand{\arraystretch}{1.12}
\begin{tabular}{@{}clccccccc@{}}
\toprule
\textbf{Comp.}
& \textbf{Method}
& \textbf{L-3.2-1B}
& \textbf{L-3.2-3B}
& \textbf{L-3.1-8B}
& \textbf{L-2-7B}
& \textbf{Q3-1.7B}
& \textbf{Q3-4B}
& \textbf{Q3-8B} \\
\midrule

\multirow{2}{*}{15\%}
& MoDeGPT
& 22.98
& \textbf{18.08}
& \textbf{14.90}
& 9.62
& \textbf{37.60}
& 27.08
& \textbf{20.76} \\
& Ours (C)
& \textbf{22.72}
& 18.15
& \textbf{14.90}
& \textbf{9.60}
& 41.94
& \textbf{26.87}
& 20.93 \\

\midrule

\multirow{2}{*}{30\%}
& MoDeGPT
& \textbf{69.56}
& 30.33
& 24.09
& \textbf{12.84}
& \textbf{56.77}
& 45.89
& 33.83 \\
& Ours (C)
& 69.83
& \textbf{30.15}
& \textbf{23.99}
& 12.87
& 59.73
& \textbf{42.27}
& \textbf{33.57} \\

\midrule

\multirow{2}{*}{40\%}
& MoDeGPT
& 102.40
& 52.10
& 38.26
& \textbf{17.83}
& 164.92
& 90.37
& 79.75 \\
& Ours (C)
& \textbf{96.20}
& \textbf{51.69}
& \textbf{38.04}
& 17.86
& \textbf{138.34}
& \textbf{74.34}
& \textbf{70.38} \\
\bottomrule
\end{tabular}
\caption{
C4 perplexity across models and compression ratios (lower is better).
Bold indicates the better result. (L) and (Q3) denote Llama and Qwen3, respectively.
}
\label{tab:c4}
\end{table*}

\subsection{Language Modeling Performance}
We analyze the performance of our corrective calibration (C) on language modeling across the different models using the WikiText-2 and C4 datasets. We compare against MoDeGPT~\citep{lin2025modegpt} given its superior language modeling capability compared to other baselines. We apply our corrective calibration (C) on top of the MoDeGPT framework, and compare against the vanilla MoDeGPT. Tables \ref{tab:wiki} and \ref{tab:c4} show the perplexity results across the various models and compression rates under WikiText-2 and C4, respectively.

\section{Inference Speedup and Memory Footprint}

Compression speedups transfer directly from UniQL and MoDeGPT since our corrections amount to a mere reordering of the pruning dimensions, rather than introducing a new compression scheme. Table \ref{tab:memory} shows the memory footprint in GB post compression.

\begin{table}[t]
\centering
\small
\setlength{\tabcolsep}{10pt}
\renewcommand{\arraystretch}{1.12}
\begin{tabular}{@{}lcccc@{}}
\toprule
\textbf{Model}
& \textbf{Base (BF16)}
& \textbf{15\%}
& \textbf{30\%}
& \textbf{40\%} \\
\midrule
Llama-3.1-8B
& 16.06
& 13.99
& 11.87
& 10.49 \\

Qwen3-4B
& 8.04
& 6.94
& 5.87
& 5.16 \\
\bottomrule
\end{tabular}
\caption{Model memory footprint in GB after modular compression.}
\label{tab:memory}
\end{table}

\section{Synergy with Post-Hoc Error Compensation Methods: EoRA } \label{appdx:eora}

We assess the synergy between our correction mechanisms and EoRA~\citep{liu2024eora}. The key distinction is that ours acts during compression itself, whereas EoRA is a post-hoc error-compensation technique that adds a parallel low-rank residual path, $R_{\mathrm{EoRA}}(\cdot)$, to compensate for weight representation errors from compression.

Since our compression hypothesis builds on SOTA modular, training-free compression frameworks, we note the following about how we experimented with EoRA:
Modular compression slices channel dimensions across a number of modules, creating a mismatch between the original and compressed weight shapes. This prevents us from directly applying EoRA per weight matrix, $\Delta W = W_{\mathrm{comp}} - W_{\mathrm{orig}}$, without adding projection matrices to the residual corrective path, introducing complexity.

To circumvent this, we apply EoRA at the block level, placing the corrective residual path across the module as a whole (MLP or attention). This leverages shape consistency at the block output and applies a module-level correction, $\Delta M = M_{\mathrm{comp}}(x) - M_{\mathrm{orig}}(x)$, resolving the shape mismatch at each block's output. It also means the block-level rank can afford to be larger than a per-weight allocation, for the same overall ratio of added parameters.

Accordingly, the corrected forward becomes $x_{\mathrm{out}} = M_{\mathrm{comp}}(x) + \eta R_{\mathrm{EoRA}}(x)$, where $\eta$ 
 is a dampening coefficient we introduce to mitigate observed effects of overcorrection from compounded correction methods (we set $\eta=0.1$). We apply EoRA with rank 256 per block after our correction-aware compression — which for example adds 88M parameters to Llama-3.2-3B model, comparable to the 97M parameters added by rank-64 EoRA in the original per-weight formulation. We consider calibration correction (C) and one round of iterative refinement (R) on the UniQL baseline at 40\% compression as shown in Table \ref{tab:eora}: 
 \begin{itemize}
\item
\textbf{EoRA further improves accuracy.} Across all configurations, 0-shot average improves by +0.41 points on average.
\item
\textbf{Ordering is preserved after EoRA.} A correction (C or C+R) favorable over the UniQL baseline before EoRA remains favorable after. That is, EoRA doesn't change which compression configuration is most favorable.
\item
\textbf{Overlapping Gains.} Our calibration corrections and EoRA target related errors at different stages, so their benefits overlap rather than fully stack.
 \end{itemize}

 \begin{table*}[t]
\centering
\small
\setlength{\tabcolsep}{5pt}
\renewcommand{\arraystretch}{1.12}
\begin{tabular}{@{}l*{4}{cc}@{}}
\toprule
& \multicolumn{2}{c}{\textbf{Llama-3.2-1B}}
& \multicolumn{2}{c}{\textbf{Llama-3.2-3B}}
& \multicolumn{2}{c}{\textbf{Llama-3.1-8B}}
& \multicolumn{2}{c}{\textbf{Qwen3-4B}} \\
\cmidrule(lr){2-3}
\cmidrule(lr){4-5}
\cmidrule(lr){6-7}
\cmidrule(l){8-9}

\textbf{Method}
& \textbf{Orig.} & \textbf{+EoRA}
& \textbf{Orig.} & \textbf{+EoRA}
& \textbf{Orig.} & \textbf{+EoRA}
& \textbf{Orig.} & \textbf{+EoRA} \\
\midrule

UniQL
& 41.37 & \textbf{41.98}
& 49.22 & \textbf{49.58}
& 56.06 & \textbf{56.53}
& 42.48 & \textbf{43.54} \\

Ours (C)
& 41.96 & \textbf{42.11}
& 49.72 & \textbf{49.86}
& 56.45 & \textbf{56.97}
& 41.99 & \textbf{42.46} \\

Ours (C+R)
& 42.12 & \textbf{42.20}
& 49.58 & \textbf{49.87}
& 56.41 & \textbf{56.81}
& 42.29 & \textbf{42.63} \\

\bottomrule
\end{tabular}
\caption{
0-shot average results for 40\% compressed models before and after EoRA.
Bold indicates the better result within each Orig./+EoRA pair.
}
\label{tab:eora}
\end{table*}

\section{Training-Free Low-Rank Compression Baselines}\label{appdx:tf-decomp}

\subsection{SVD-LLM Whitening-based Single Weight Decomposition}\label{appdx:svdllm}

SVD-LLM~\citep{wang2025svdllm} leverages an activation-aware whitening approach to decompose each weight matrix $W$ into its singular matrices ($U$ and $V^T$) and singular values $\Sigma$. The approach involves computing the covariance matrix at the input to each weight matrix; deriving a whitening matrix $S$; and performing $\mathrm{SVD}(WS)$ to obtain singular directions that minimize the output reconstruction error. Whitening enables identifying the most important weight directions with respect to the observed input distribution.

\subsection{MoDeGPT Joint Low-Rank Decomposition}\label{appdx:modegpt}
MoDeGPT~\citep{lin2025modegpt} proposes to apply decomposition to multiple matrices jointly as modules. They define three functional modules (MLP, QK, and VO), and propose a customized decomposition operation for each tailored to the number of nonlinearities within each module. The decomposition for the three module groupings are as follows: 

\begin{itemize}
    \item \textbf{MLP.} Nyström approximation is used to select the most important intermediate dimensions based on the activation correlation matrix collected at the down projection input. The same selected dimensions are retained across the up and gate projections, while the down projection is adjusted to reconstruct the original MLP output.

    \item \textbf{QK.} Key--Query compression is performed independently for each attention head. Since the query and key projections jointly determine the attention scores, MoDeGPT selects a shared subset of dimensions based on their joint activation statistics.

    \item \textbf{VO.} A two-stage SVD is used to jointly compress the value and output projections. The value projection is first decomposed using its input activation correlation, before the output projection is decomposed while accounting for the reduced value representation. The obtained decompositions are then combined to approximate the complete VO module.
\end{itemize}

Though MoDeGPT derives module-specific reconstruction procedures with theoretical guarantees, the pseudo-inverse operation in its MLP decomposition introduces a computational bottleneck, requiring the collection and retention of activation-correlation statistics and higher-precision computation for numerical stability.

\subsection{UniQL Joint Low-Rank Compression}\label{appdx:uniql}

Though UniQL~\citep{chiang2026uniql} is similar to MoDeGPT, it presents an alternative pseudo-inverse-free joint low-rank compression solution that offers considerably faster compression times, as shown in Table~\ref{tab:wallclock}. Specifically, UniQL replaces the expensive Nyström approximation used for the MLP module with an efficient column-sorting approach. The intermediate dimensions are ranked by their importance scores and rearranged such that any target number of dimensions can be retained through direct truncation. Briefly, the module decompositions can be summarized as follows:

\begin{itemize}
    \item \textbf{MLP.} A structured sorting matrix is derived from the intermediate activation correlation matrix collected prior to the down projection. The same ordering is applied consistently across the up, gate, and down projections, placing the most important intermediate dimensions first and allowing the remaining dimensions to be directly truncated for any target rank.

    \item \textbf{QK.} Query and key dimensions are ranked based on their joint activation statistics prior to the attention calculation. The resulting shared ordering is applied independently to each attention head along the head dimension, allowing the least important shared dimensions to be directly truncated.

    \item \textbf{VO.} The input covariance matrix is collected for each value head and used within a two-stage SVD decomposition across the value and output projections. This jointly reduces the value representation while accounting for its interaction with the corresponding output projection.
\end{itemize}

It should be noted that UniQL is a system-aware solution and performs its structural truncation in multiples of: (a) 16 for the attention weights; and (b) 128 for the MLP weights.

\section{Limitations}

Our approach remains to be evaluated in other settings: (1) Extensive testing on larger model sizes (>8B models); (2) Aggressive compression ratios (<50\%);  (3) Emerging architectures such as hybrid and state space models with different compositions. In principle, our methods are designed to be scalable to any model size and type. Still, we stress that rigorous assessment in these scenarios remains understudied.

Our evaluations leverage BI scores for layer sensitivity estimation following recent state-of-the-art practice. However, the resulting rank allocation remains conditioned on the selected scoring mechanism. Investigating the robustness of our findings across different sensitivity metrics remains a topic for future investigation.

\end{document}